\documentclass{ieeeaccess}

\usepackage{cite}
\usepackage{url}
\usepackage{amsmath,amssymb,amsfonts,subfig}
\usepackage{algorithmic}
\usepackage{graphicx}
\usepackage{textcomp}

\graphicspath{{./pdf/}}
\usepackage{color}
\usepackage{bm}
\usepackage{booktabs}

\def\BibTeX{{\rm B\kern-.05em{\sc i\kern-.025em b}\kern-.08em
    T\kern-.1667em\lower.7ex\hbox{E}\kern-.125emX}}
\begin{document}

\title{Development of the Lifelike Head Unit for a Humanoid Cybernetic Avatar `Yui' and Its Operation Interface}
\author{
\uppercase{Mizuki Nakajima}\authorrefmark{1}, 
\uppercase{Kaoruko Shinkawa}\authorrefmark{2}, 
and \uppercase{Yoshihiro Nakata}.\authorrefmark{2}, 
}
\address[1]{Tokyo Denki University, Adachi-ku, Tokyo 120-8551, Japan}
\address[2]{The University of Electro-Communications, Chofu, Tokyo, 182-8585, Japan}
\tfootnote{``This work was supported by JST Moonshot R\&D Grant Number JPMJMS2011.''}

\markboth
{Mizuki Nakajima \headeretal: Development of the Lifelike Head Unit for a Humanoid Cybernetic Avatar `Yui' and Its Operation Interface}
{Mizuki Nakajima \headeretal: Development of the Lifelike Head Unit for a Humanoid Cybernetic Avatar `Yui' and Its Operation Interface}

\corresp{Corresponding author: Mizuki Nakajima (e-mail: mizuki.nakajima@mail.dendai.ac.jp).}

\begin{abstract}
In the context of avatar-mediated communication, it is crucial for the face-to-face interlocutor to sense the operator's presence and emotions via the avatar. Although androids resembling humans have been developed to convey presence through appearance and movement, few studies have prioritized deepening the communication experience for both operator and interlocutor using android robot as an avatar. Addressing this gap, we introduce the ``Cybernetic Avatar `Yui','' featuring a human-like head unit with 28 degrees of freedom, capable of expressing gaze, facial emotions, and speech-related mouth movements. Through an eye-tracking unit in a Head-Mounted Display (HMD) and degrees of freedom on both eyes of Yui, operators can control the avatar's gaze naturally. Additionally, microphones embedded in Yui's ears allow operators to hear surrounding sounds in three dimensions, enabling them to discern the direction of calls based solely on auditory information. An HMD's face-tracking unit synchronizes the avatar's facial movements with those of the operator. This immersive interface, coupled with Yui's human-like appearance, enables real-time emotion transmission and communication, enhancing the sense of presence for both parties. Our experiments demonstrate Yui's facial expression capabilities, and validate the system's efficacy through teleoperation trials, suggesting potential advancements in avatar technology.
\end{abstract}


\titlepgskip=-15pt

\maketitle

\section{Introduction}
Teleoperated robot (avatar) technologies have been studied mainly to improve operability and the sense of presence obtained by the operator in remote works \cite{teleope}.
Among these studies, avatar operation systems using immersive interfaces have recently attracted attention.
Interfaces for immersive experiences, such as Head-Mounted Display (HMD) and Cave automatic virtual environment (CAVE) \cite{cave} that change presented images response to the operator's movements or Torus treadmill \cite{treadmill} that converts the operator's walking or running motion to input, are known to have advantages in terms of presence and operability compared to general interfaces such as a desktop monitor or a joystick \cite{hmd_emo_comp,hmd_presense_comp,hmd_vs_cave,hmd_vs_desktop,hmd_att_comp}. 
In the literature \cite{icub3}, the operator's movements are reproduced on an avatar using an HMD and a treadmill.
The avatar and its operation system have been verified through demonstrations in which the avatar is controlled from a remote location more than 290 km away.
Tachi et al. developed TELESAR VI, an avatar with movable full-body, and a system to control the avatar through HMDs and motion trackers \cite{TELESAR_vi}.
The delicate works in remote locations were realized by feeding back the temperature and pressure of the avatar's fingertips to the operator.
The literature \cite{hmd_nao} investigated the experiences of both operators and interlocutors in cooperative work with a person at a distance and an avatar controlled by the operator using HMD and motion tracker.
It was suggested that the subjective experience of the operator was enhanced as the operability of the avatar was improved.
In the literature \cite{hmd_modality}, the influence of the operator's interface on the avatar's presence is investigated using Daryl \cite{daryl}, an avatar with an eye camera and head  movement.

In contrast, in human interactions, the conveyance of emotions through facial expressions is suggested to be important for the communication \cite{Mehrabian2017-tz}. 
Therefore, androids with human-like appearances which can represent internal states through modalities such as gaze and facial expressions have been developed.
An android Nikola \cite{Nikola} has 35 degrees of freedom (DoFs) for facial skin and head movements.
Nikola can express six basic emotions such as Anger, Disgust, Fear, Happiness, Sadness, and Surprise, by using these .
Furthermore, the relationship between the speed of changing facial expressions and the perceived naturalness by the interlocutor is discussed in \cite{Nikola}.
In the literature \cite{hrp4c}, the android HRP-4C which can walk with two legs is developed.
HRP-4C is developed by referring to the body shape of an adult Japanese woman and features a human-like head unit.
ERICA \cite{erica} has been developed as an interactive autonomous android. 
ERICA possesses a human-like appearance and is capable of autonomous dialogue and gestures synchronized with speech. 
Some androids utilize displays or screens for the facial area to express emotions \cite{socibot, Furhat}.
They can express emotions by projecting simplified facial expressions or facial images of real humans.
Even in companies such as Engineered Arts Ltd., androids with human-like appearance have been developed, suggesting a high societal interest in the use of androids \cite{ameca}.

Many of these existing studies have focused on either the operator or the interlocutor, and have not emphasized the enhancing of the communication experience for both the operator and the interlocutor when they interact through the avatar.
In recent years, devices such as Vive Pro Eye, FOVE 0, PICO Neo3 Pro Eye, and Meta Quest Pro have been released that can acquire the user's facial expressions while using an immersive interface.
The iCub3 \cite{icub3} uses a Vive Pro Eye to reproduce facial expressions on the avatar while presenting video through the HMD.
However, the avatar's facial expression is very simplified and expressed by switching LEDs placed at the mouth and eyebrow positions.
It has been reported that human facial expressions are expressed by a combination of multiple facial movements called Action Units \cite{action_unit}, and the reproduction of facial expressions by iCub3 is not sufficient.

\begin{figure}[t]
  \begin{center}
    \resizebox*{8.2cm}{!}{\includegraphics{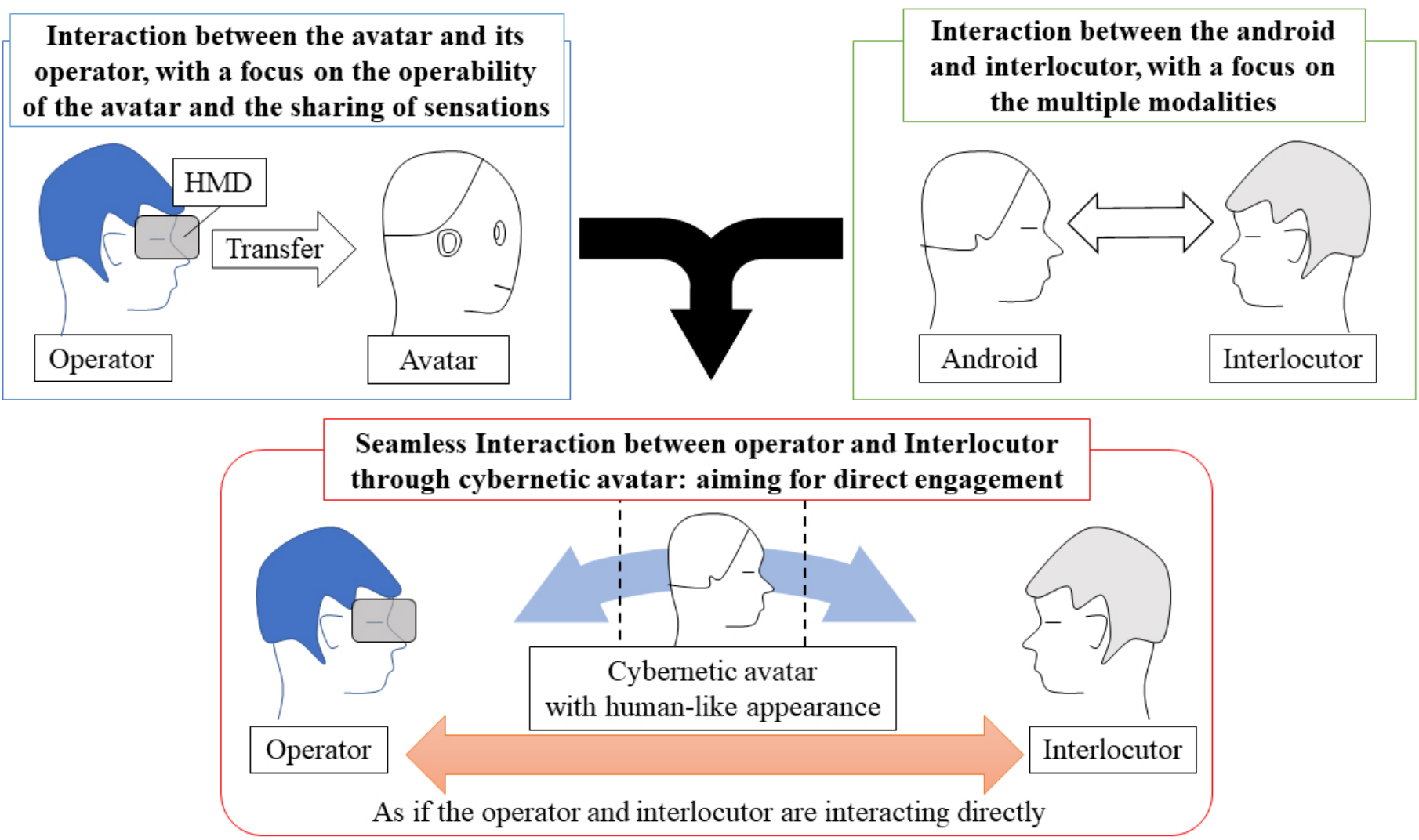}}
    \caption{Positioning of the research}
    \label{fig:positioning}
  \end{center}
\end{figure}

Based on this background, this paper develops a head unit of Cybernetic avatar \cite{cybernetic_avatar} `Yui' for the main purpose of remote communication, and a system to control CA through HMD, which is an immersive interface.
Fig.~\ref{fig:positioning} shows the positioning of this study.
We aim to enhance the interaction between the operator and the interlocutor by enhancing both the interaction between the operator and the avatar, and between the avatar and the interlocutor.
The head unit has a human-like appearance and can express changes in head posture, gaze, facial expressions, and mouth movements associated with speech using 28 facial actuation points.
The eye tracking unit and face tracking unit mounted on the HMD can acquire changes in the operator's eye movements and facial expressions.
The operator's behavior can be represented on the avatar by synchronizing their motion.
The camera and microphones are mounted on the avatar's eyes and ears, and the operator can control the avatar while perceiving three-dimensional vision and sound by using them.
The developed avatar can enhance interaction between the operator and the avatar by providing the information about the distance and positional relationship between the avatar and the surroundings through stereo sounds and stereo images.
The interaction between the avatar and the interlocutor can be enhanced by providing non-verbal behaviors such as the operator's facial expressions and eye movements to the interlocutor through the avatar, in addition to the operator's voice.
The main contribution of this study is follows:
\begin{itemize}
    \item Development of a system that focuses on the presence of the avatar and the sense of presence of both the operator and the interlocutor by using an immersive interface capable of detecting facial expressions and eye movements.
    \item Development of an avatar head unit that can reproduce 7 main facial expressions, eye movement, and head orientation using 28 facial actuation points.
    \item Implementation and simple verification of the presentation system of a stereophonic image and stereophonic sound by using cameras mounted in both eyes and microphones mounted in both ears.
    \item Implementation and simple verification of a system that reproduces the operator's facial expressions on the avatar.
\end{itemize}
The cybernetic avatar `Yui' is a successor to `ibuki' \cite{ibuki}, a mobile android avatar developed by the authors in their previous research, and has been improved in many areas such as freedom of head movement, quietness, and viewing angle.
The avatar system developed in this study suggests the possibility of improving communication experience for both the operator and the interlocutor.
Further research and practical application of this technology is expected to lead to the diffusion and application of avatar technology.

\section{System overview}
\begin{figure}[t]
  \begin{center}
    \resizebox*{8.2cm}{!}{\includegraphics{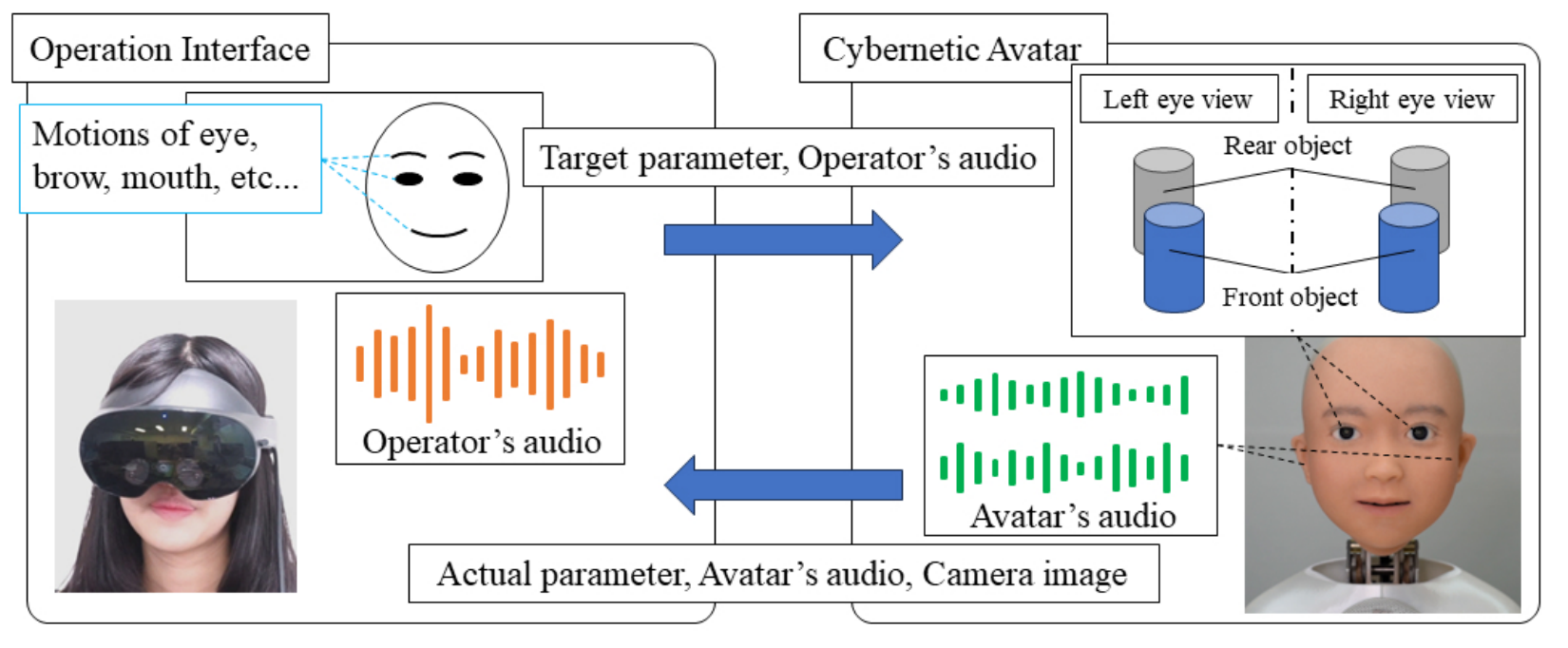}}
    \caption{System overview}
    \label{fig:system_overview}
  \end{center}
\end{figure}

Fig.~\ref{fig:system_overview} shows the overview of the developed system.
The system consists of the Cybernetic Avatar (CA) and the Operation Interface, and each part communicates with each other to operate.
The CA and Operation Interface operate independently, and are able to communicate with each other by connecting to the same network.
In the empirical verification of this study, the system can connect the CA at remote locations by constructing Virtual Private Network (VPN).
In this paper, the network is constructed by connecting directly via a wired LAN.
The CA has various perceptions and a head that can change its facial expression and posture, and can transmit acquired information or receive instructions through the Operation Interface.
The Operation Interface has the function of presenting information received from the CA to the operator and transmitting the operator's instructions to the CA.
In the following sections, the details of the CA and the Operation Interface will be introduced.

\section{Cybernetic Avatar}
\begin{figure}[t]
  \begin{center}
    \resizebox*{8.2cm}{!}{\includegraphics{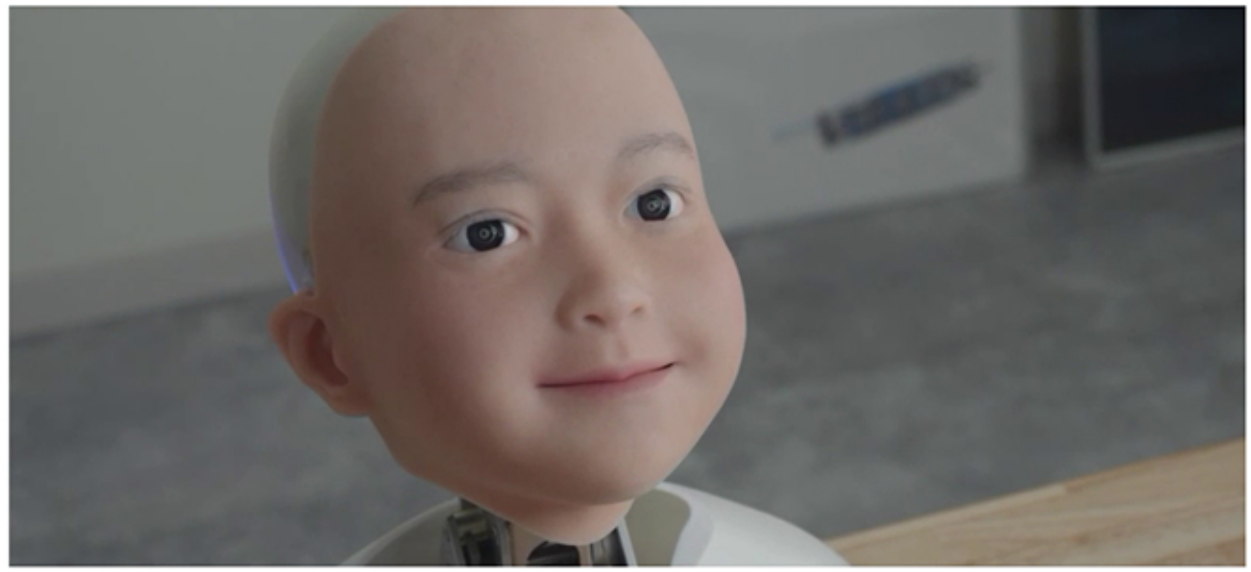}}
    \caption{Cybernetic Avatar `Yui'}
    \label{fig:yui}
  \end{center}
\end{figure}

Fig.~\ref{fig:yui} shows an appearance of the head unit of the CA.
This study assumes no specific operators and interlocutors.
Therefore, it is important that the design be familiar to as many people as possible.
Aiming to achieve communication with as broad an audience as possible, the CA is designed as a child-like and neutral appearance.
In addition, by designing the CA with as few characteristics as possible, as in the case of Telenoid \cite{telenoid, telenoid_with_older}, we aim to evoke the operator through  the CA's behavior and facial expressions.
By reminding the operator to the interlocutor through the CA's behavior, it is expected to improve the communication experience between the operator and the interlocutor, regardless of the differences in appearance between the CA and the operator.
This section introduces the functions of the CA.

\subsection{Mechanical design}
\begin{figure}[t]
  \begin{center}
    \subfloat[Degrees of freedom of the neck part]{
    \resizebox*{2.3cm}{!}{\includegraphics{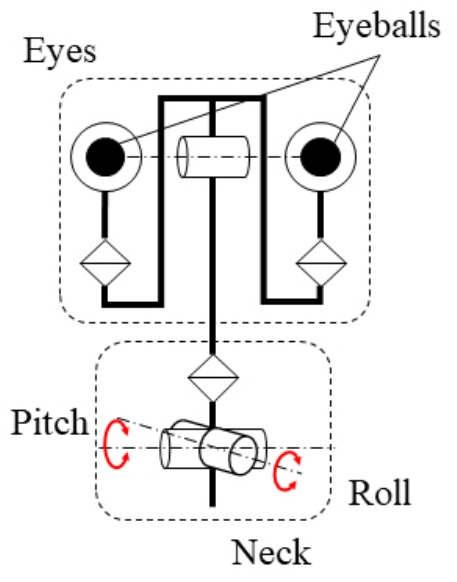}}
    \label{fig:neck_1}}
    \subfloat[Model of the neck part]{
    \resizebox*{5.8cm}{!}{\includegraphics{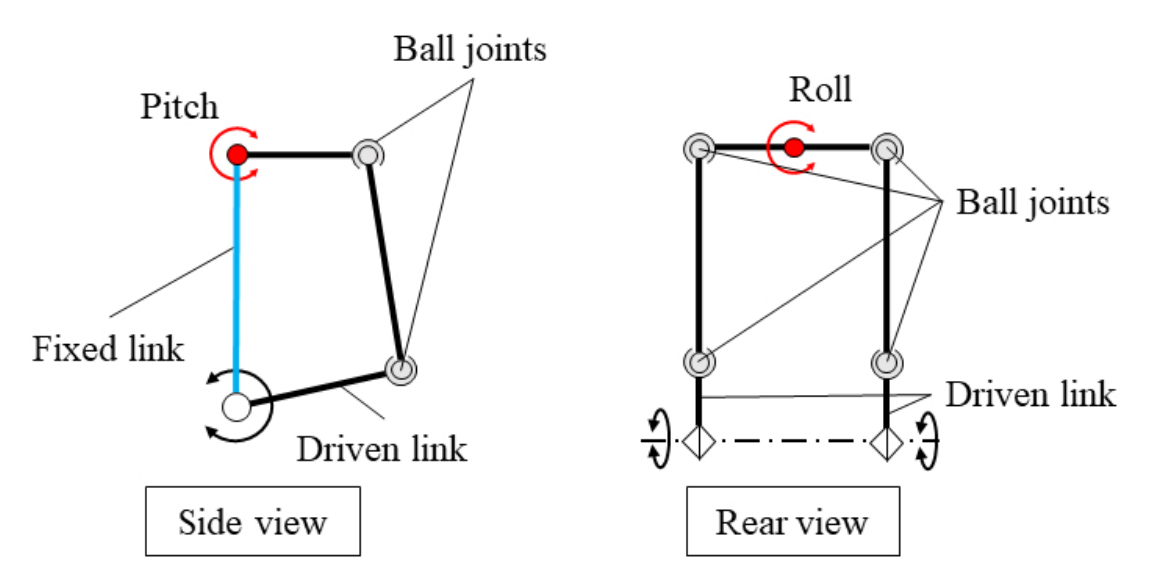}}
    \label{fig:neck_2}}
    \caption{Outline of the neck part}
    \label{fig:neck}
  \end{center}
\end{figure}

This subsection will describe the mechanical design of the head unit.
Furthermore, the mechanical design of the facial actuation will be described in Sec.~\ref{sec:facial_act}.
This robot has 21 degrees of freedom, with some motors directly driving joints and others utilizing both forward and reverse rotations to actuate two different points by pulling wires. 
This unique assignment of motor rotations increases the total number of actuation points, resulting in 28 actuation points.
We use electric motors for the actuator in consideration of quietness and responsiveness.
Fig.~\ref{fig:neck} shows the actuation point arrangement of the neck and eyes, and the model of the neck part.
The left and right eyeballs rotate independently around the yaw axis and simultaneously around the pitch axis.
The actuation point of the neck is arranged so that each of the roll-pitch-yaw axes is orthogonal at a single point.
The pitch and yaw axes of the neck operate in concert via a differential linkage mechanism. 
The average of the rotation of the two axes is the rotation around the pitch axis, and the difference of the rotation of the two axes is the rotation around the roll axis.
The use of a differential linkage mechanism allows operation with greater torque than a single actuator for each axis, and arrangement of the neck actuator in the body part.
The weight of the head above the neck is approximately 2.0 kg, which is greater than other structures of the head unit.
In addition, a lot of actuators for facial expression movements are placed inside of the head unit  and there is not enough space for placing the neck actuators.
Therefore, we use the differential linkage mechanism for the neck part.

\subsection{Facial actuation}
\label{sec:facial_act}
\begin{figure}[t]
  \begin{center}
    \resizebox*{8.2cm}{!}{\includegraphics{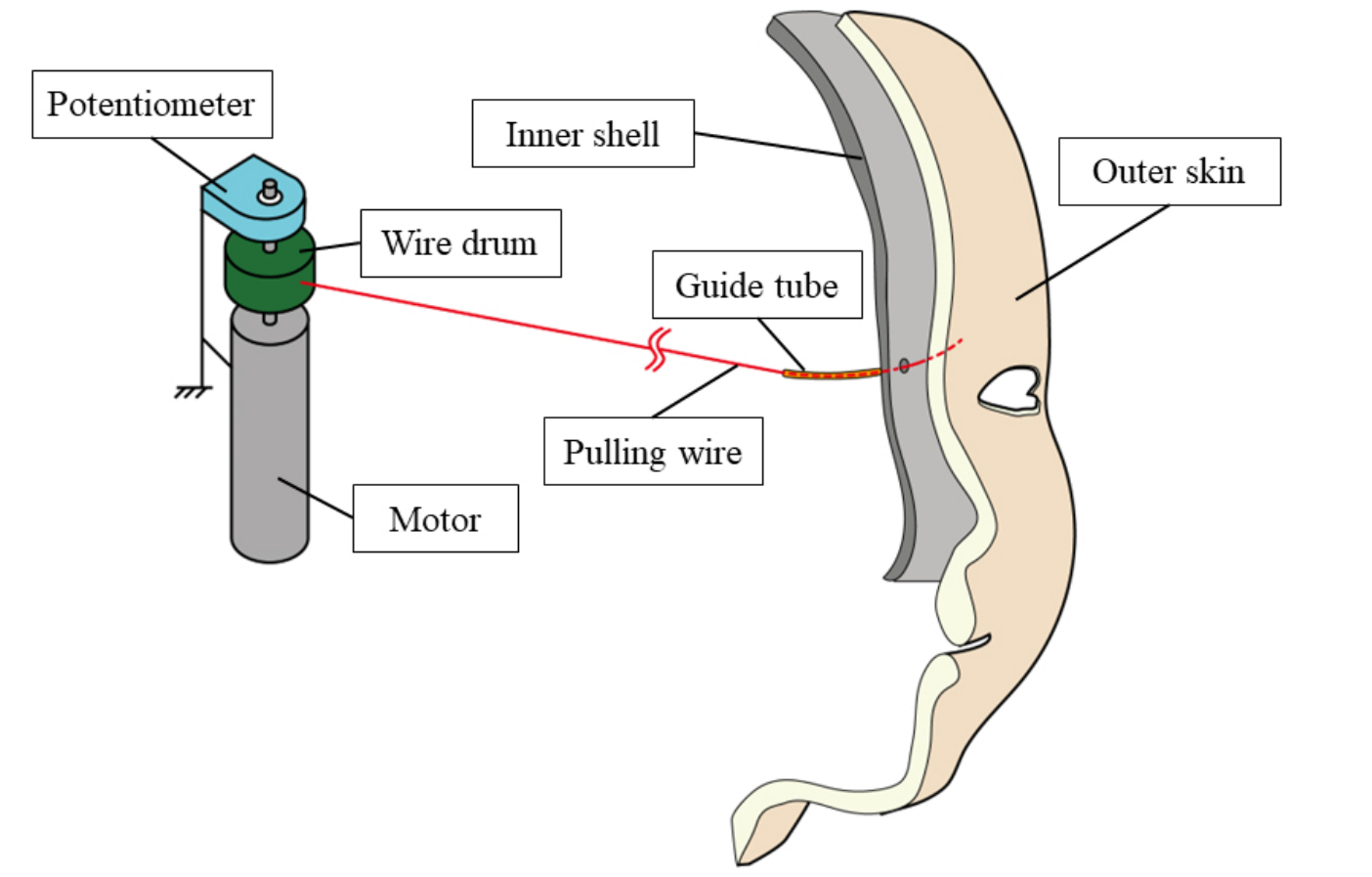}}
    \caption{Mechanical design for facial actuation}
    \label{fig:facial_mech}
  \end{center}
\end{figure}

\begin{figure}[t]
  \begin{center}
    \resizebox*{8.2cm}{!}{\includegraphics{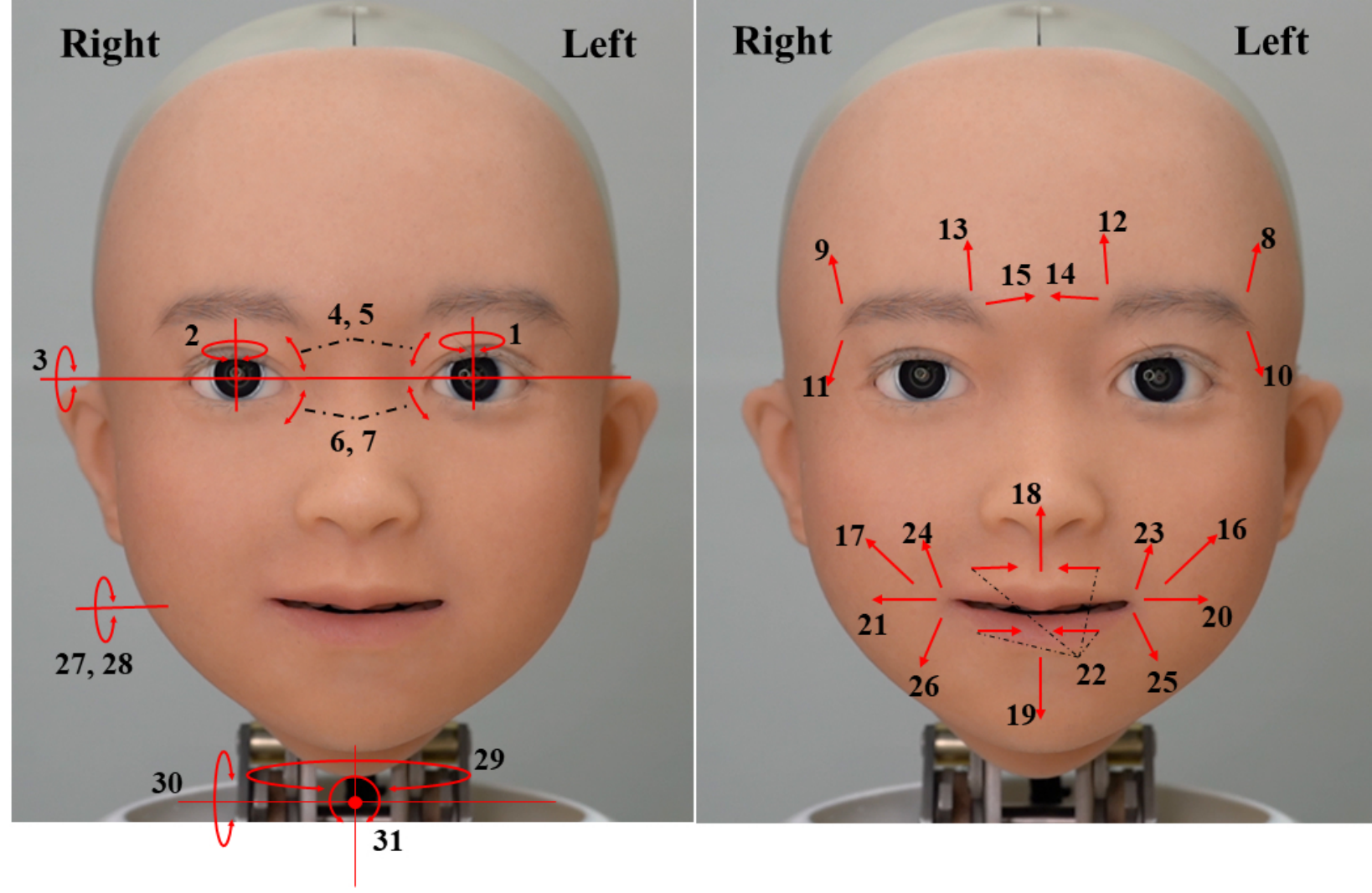}}
    \caption{Facial actuation point (Left: points controlled by rigid links, Right: points controlled by pulling wire)}
    \label{fig:facial_actuation_point}
  \end{center}
\end{figure}

\begin{figure}[t]
  \begin{center}
    \resizebox*{8.2cm}{!}{\includegraphics{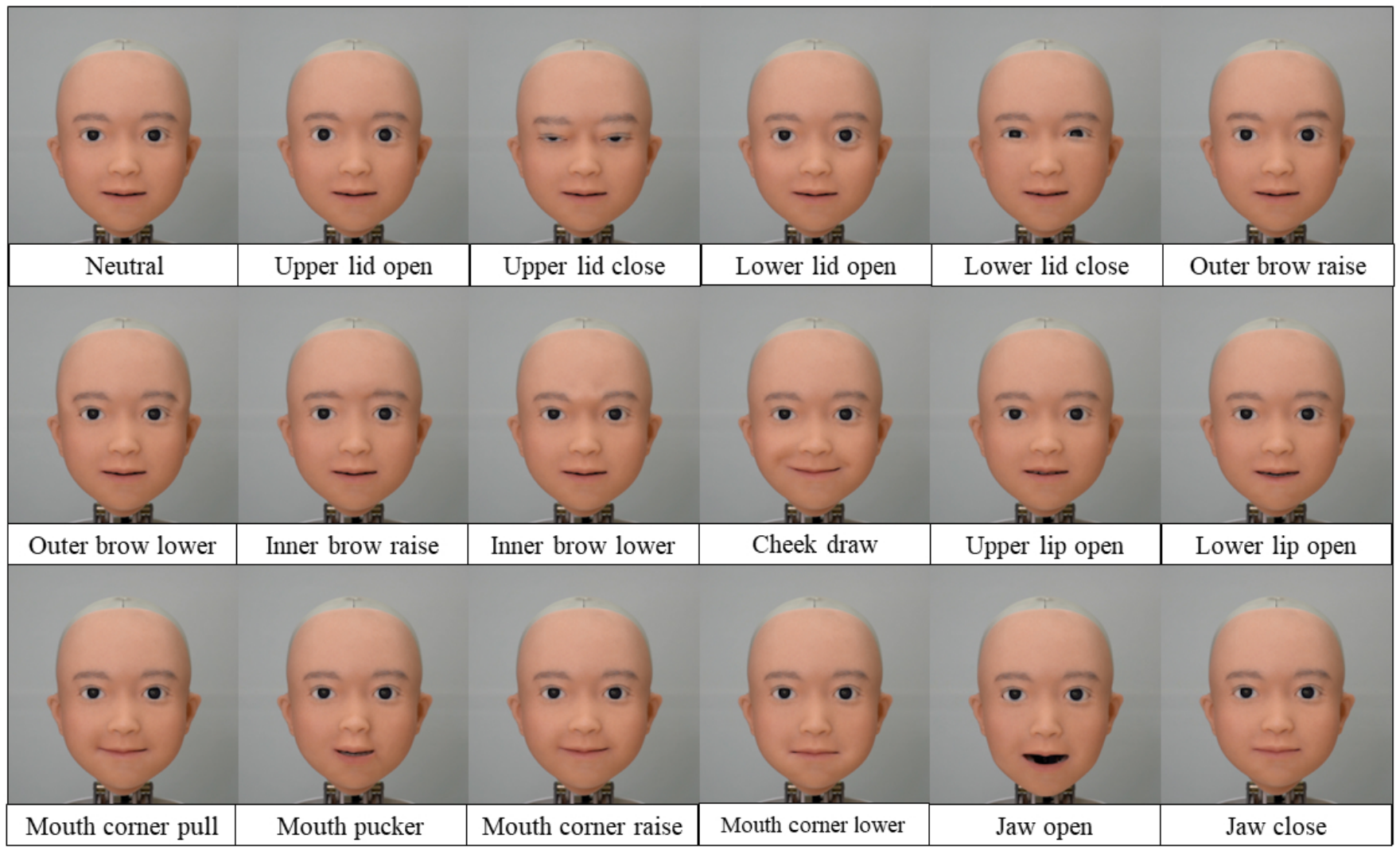}}
    \caption{Range of motion}
    \label{fig:motion_limit}
  \end{center}
\end{figure}

Fig.~\ref{fig:facial_mech} shows the details of the mechanical design for the facial actuation.
The ends of wires are attached to the underside of the skin and the actuator for facial actuation through wire guide tubes placed in the head shell.
The wire drum is attached to the actuator, and the facial skin moves by winding up the wire using the drum.
The facial expression changes from a neutral state when the skin is pulled by the wire, and returns to a neutral expression by the elasticity of the skin when the wire is loosened.
Some actuators move 2 actuation points with a single actuator by pulling different wires in forward and reverse rotation.
In addition, the parts that require higher torque and faster response than facial expressions, such as eyelids, eye orientation, jaw opening and closing, and head orientation, are driven from the output of the actuator via rigid links.
Table~\ref{tab:head_motion_act} and Fig.~\ref{fig:facial_actuation_point} show the overview of each head actuation points.
Table~\ref{tab:motion_range} and Fig.~\ref{fig:motion_limit} show the motion ranges of the eye and neck, and of the other actuation points, respectively.
In Fig.~\ref{fig:motion_limit}, the actuation points placed symmetrical are moved simultaneously.
Furthermore, in relation to the Action Unit described below, the opening and closing of the eyelids and the jaw are divided into 2 actuation points in Table~\ref{tab:head_motion_act}.

It is known that human facial expressions can be expressed by a combination of facial deformations called Action units \cite{action_unit}.
The arrangements of the actuation point in the face were decided based on the Action units.
For structural reasons, some Action units are realized by multiple head actuation points.
Tables~\ref{tab:emotions_and_au} shows the Action units and the head actuation points that are controlled in each emotional expression.
The descriptions of the Action unit and the relationships between the Action units and the head actuation points are shown in Tables~\ref{tab:dscrpt_au} and \ref{tab:au_and_mot}, respectively.
These facial expressions are reproduced on the CA by driving the corresponding actuator.
The head actuation points 10 and 11 do not have a directly corresponding Action unit, but are used to emphasize Action unit 4.
In contrast, for Action unit 9, there is no corresponding head actuation point.
However, a close facial expression is produced by moving the head actuation points 7, 14 and 15 simultaneously.
Therefore, Action unit 9 is reproduced by using these actuation points simultaneously in this paper.
Similarly, the Action unit 17 is substituted by actuation point 28.
In addition, Action unit 1 and 4 are used for expressing Fear emotion, but the CA can not move these simultaneously.
This is because each motion is assigned to forward and reverse rotation of a single motor.
Thus, Action unit 4 is not used for expressing Fear emotion in this paper.

\begin{table}[t]
    \caption{Relationship between head motion and actuators}
    \centering
    \begin{tabular}{ccc}
         \toprule
         Head motion No. &   Actuator No.  &   Description \\ \hline
         1 &   1  &  Left eye left and right\\
         2 &   2  &  Right eye left and right\\
         3 &   3  &  Eye up and down\\
         4 &   4+  &  Upper eyelid open\\
         5 &   4-  &  Upper eyelid close\\
         6 &   5+  &  Lower eyelid open\\
         7 &   5-  &  Lower eyelid close\\         
         8 &   6+  &  Left outer eyebrow up\\
         9 &   7+  &  Right outer eyebrow up\\
         10 &  6-  &  Left outer eyebrow down\\
         11 &  7-  &  Right outer eyebrow down\\
         12 &  8+   & Left inner eyebrow up\\
         13 &  9+   & Right inner eyebrow up\\
         14 &  9-  &  Left inner eyebrow frown\\
         15 &  8-   & Right inner eyebrow frown\\
         16 &  10   & Left check pull \\
         17 &  11   &  Right check pull\\
         18 &  12  &  Upper lip up\\
         19 &  13   &  Lower lip down\\
         20 &  14+   &  Left corner of the mouth pull \\
         21 &  15+   &  Right corner of the mouth pull\\
         22 &  14- and 15-   & Mouth pucker\\
         23 &  16+   &  Left corner of the mouth up\\
         24 &  17+  &  Right corner of the mouth up\\
         25 &  16-   &  Left corner of the mouth down\\
         26 &  17-   &  Right corner of the mouth down\\
         27 &  18+  &  Jaw clench\\
         28 &  18-   &  Jaw drop\\
         29 &  19   &  Head turn\\
         30 &  20+/- and 21+/-   &  Head up and down\\
         31 &  20+/- and 21-/+  &  Head tilt\\ \hline
    \end{tabular}    
    \label{tab:head_motion_act}
\end{table}

\begin{table}[t]
    \caption{Motion range of actuators}
    \centering
    \begin{tabular}{ccc}
         \toprule
         Head motion No. & Description &  Range [deg.]\\ \hline
         1   &  Left eye turns left/right  & -35 -- 35\\
         2   &  Right eye turns left/right  & -35 -- 35\\
         3   &  Eyes up/down  & -14 -- 8\\
         29  &  Head turn & -83 -- 83\\
         30  &  Head up and down & -30 -- 40\\
         31  &  Head tilt &  -21 -- 21 \\ \hline
    \end{tabular}    
    \label{tab:motion_range}
\end{table}

\begin{table}[t]
    \caption{Relationship between action units and emotions}
    \centering
    \begin{tabular}{ccc}
         \toprule
         Emotion    &  Action unit                  & Head motion No.\\ \hline
         Happiness  & 6, 12                         & 16, 17, 23, 24\\
         Sadness    & 1, (4), 15, (17)              & 10--15, 25--26, 28\\
         Surprise   & 1, 2, 5, 25 or 26             & 4, 8, 9, 12, 13, 18, 19, 27\\
         Fear       & 1, 2, 4, 5, 7, 20, (25 or 26) & 4, 5, 7--9, 12--15, 18--21, 25--27\\
         Anger      & 4, 5 and/or 7, 22, 23, 24     & 5, 7, 14, 15, 18, 19, 22, 28\\
         Disgust    & 9 and/or 10, (25 or 26)       & 7, 14, 15, 18, 19, 27\\
         Contempt   & Unilateral 12, unilateral 14  & 20, 21, 23, 24 \\ \hline
    \end{tabular}    
    \label{tab:emotions_and_au}
\end{table}

\begin{table}[t]
    \caption{Descriptions of action units (AU)}
    \centering
    \begin{tabular}{cccc}
         \toprule
         AU &   Description &   AU  &   Description\\ \hline
         1 &   Inner brow raiser &   14  &   Dimpler\\
         2 &   Outer brow raiser &   15  &   Lip corner depressor\\
         4 &   Brow lowerer &   17  &   Chin raiser\\
         5 &   Upper lid raiser &   20  &   Lip stretcher\\
         6 &   Cheek raiser &   22  &   Lip funneler\\
         7 &   Lid tightener &   23  &   Lip tightener\\
         9 &   Nose wrinkler &   24  &   Lip pressor\\
         10 &   Upper lip raiser &   25  &   Lips part\\
         12 &   Lip corner puller &   26  &   Jaw drop\\ \hline
    \end{tabular}    
    \label{tab:dscrpt_au}
\end{table}

\begin{table}[t]
    \caption{Relationship between action units and head motion}
    \centering
    \begin{tabular}{ccc}
         \toprule
         AU &   Description &   Head motion No.\\ \hline
         1 &   Inner brow raiser &   12, 13  \\
         2 &   Outer brow raiser &   8, 9  \\
         4 &   Brow lowerer &   14, 15 \\
         5 &   Upper lid raiser &   4  \\
         6 &   Cheek raiser &   16, 17  \\
         7 &   Lid tightener &   5, 7  \\
         9 &   Nose wrinkler &   (7, 14, 15)  \\
         10 &   Upper lip raiser &   18  \\
         12 &   Lip corner puller &   23, 24 \\ 
         14  &   Dimpler &   20, 21 \\
         15  &   Lip corner depressor &   25, 26 \\
         17  &   Chin raiser &   (28) \\
         20  &   Lip stretcher &   20, 21, 25, 26 \\
         22  &   Lip funneler &   18, 19, 22 \\
         23  &   Lip tightener &   22 \\
         24  &   Lip pressor &   28 \\
         25  &   Lips part &   18, 19 \\
         26  &   Jaw drop &   27 \\ \hline
        \end{tabular}    
    \label{tab:au_and_mot}
\end{table}

\subsection{Perception}
\begin{figure}[t]
  \begin{center}
    \resizebox*{8.2cm}{!}{\includegraphics{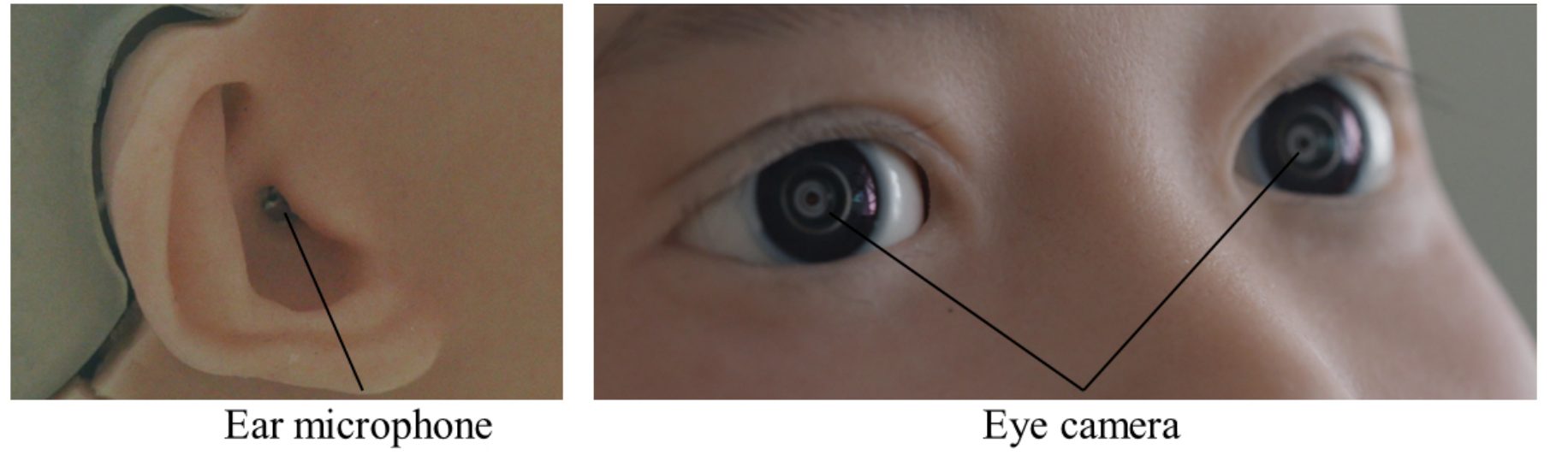}}    
    \caption{Enlarged view of the eye and ear}
    \label{fig:ear_and_eye}
  \end{center}
\end{figure}

The CA has cameras and microphones as the sensory organs for acquiring the information about the interlocutor and surroundings.
The CA presents this information to the operator through the Operation Interface, and shares the information about the surroundings in which the CA exists.
Fig.~\ref{fig:ear_and_eye} shows the microphones and cameras attached to the ear and eye of the CA, respectively.

\subsubsection{Visual Perception}
The CA has two wide-angle cameras in left and right eye positions, and can acquire
the images viewed from these positions.
These images contain the parallax that is close to it that humans achieve with their own eyes.
Humans recognize the distance to objects and the relative relationship between objects by the parallax.
The CA may be able to recognize the distance and relative relationship by the parallax obtained by the cameras similarly.
In addition, the CA can rotate the eyeballs in 2 directions.
The center of the angle of view of the camera mounted on the eyeball changes accordingly the rotation of the eyeball.
Due to the characteristics of wide-angle cameras, the image at the center of the angle of view is the clearest with the least distortion.
The camera can present a clearer image to the operator than when the same camera is used in a fixed rotation by pointing the camera in the desired direction along with eyeball movements.
In both humans and the android avatar, the eyeball is clearly lighter than the whole head.
Therefore, the eyeball motion is clearly superior in terms of energy requirement and response speed compared to the whole head motion.
The literature \cite{gaze_movement} reported that synchronization of eye movements between operators and avatars improves the operability of the robot and reduces the workload.
The field of view of the cameras attached to the Yui and motion range of the eyeballs in left and right direction are larger than the android ibuki \cite{ibuki}, which is used in the literature \cite{gaze_movement}.
Thus, the system can follow the actual eye movements more closely and is expected to improve spatial presence.

\subsubsection{Audio Perception}
The microphones are placed in each ear of the CA.
The microphones are embedded outward at a position corresponding to a human's ear canal, enabling sound acquisition at a position almost the same as that of a human.
In addition, the auricle-like structure of the CA provides a sound collecting effect like the human's auricle.
The microphone is a CS-10EM (Roland Corp.), which is capable of acquiring stereo audio from the left and right microphones.

\subsection{Electrical system}
\begin{figure}[t]
  \begin{center}    
    \resizebox*{8.2cm}{!}{\includegraphics{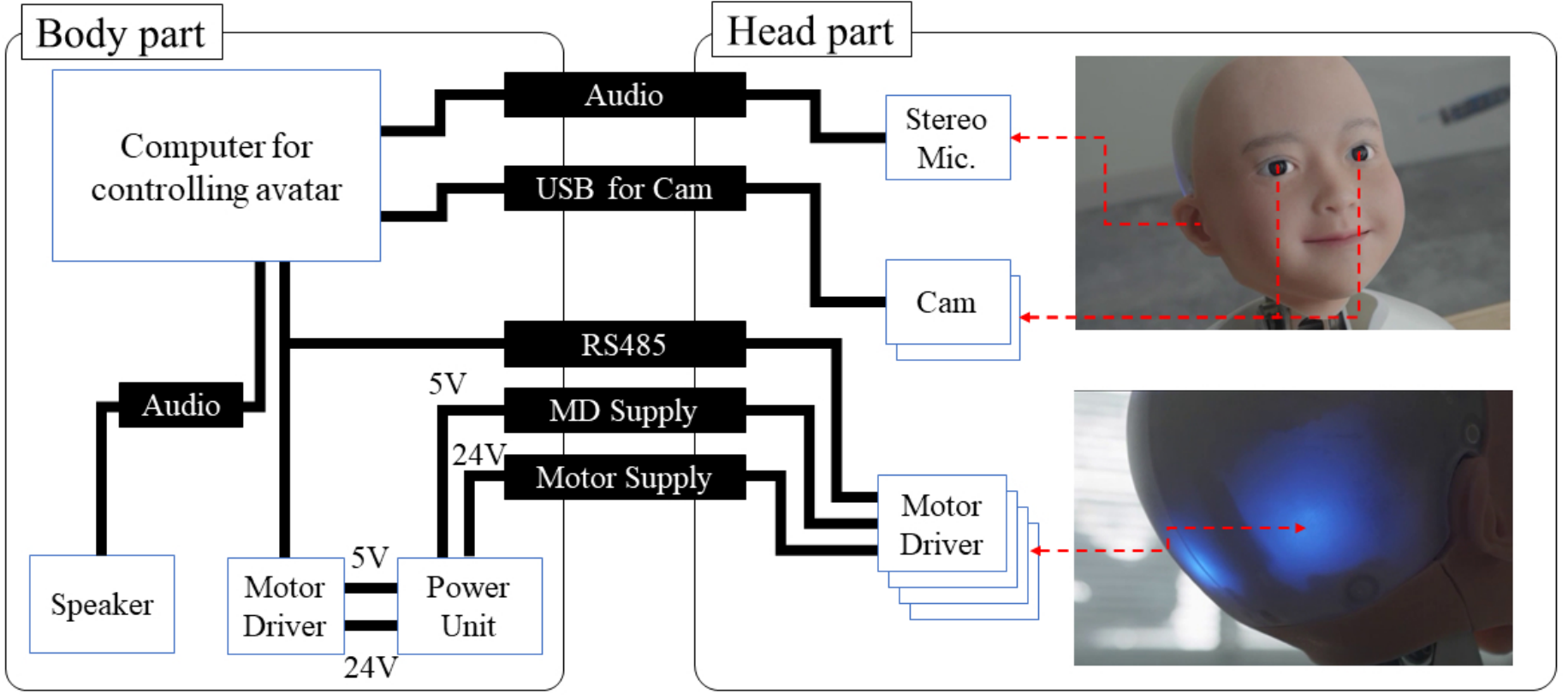}}    
    \caption{Overview of the electrical system}
    \label{fig:elec_sys}
  \end{center}
\end{figure}

Fig.~\ref{fig:elec_sys} shows the electrical connection of the system.
The microphones and cameras, motors and its driver board are mounted inside the head unit.
These components are connected to the computer and the power unit mounted in the body part of the CA.
Due to the internal space of the head unit, the motor for the neck is connected to a driver board mounted on the body part.
The driver board has five connectors to connect the motor and potentiometer and can control five motors simultaneously.
Each connected motor is individually controllable and controlled to different angles.
Four driver boards are mounted inside the head unit for controlling 19 motors excluding the 2 motors for the neck movement.
The driver boards connect to the computer via RS485 serial communication.
In addition, each driver board can connect to adjacent boards by daisy-chain connection, contributing to a reduction in wiring space.

\subsection{Software and Control}
This subsection presents a software framework for transmitting the camera and microphone information to the operator and a framework for reproducing the audio information transmitted from the operator and the operator's facial expressions on the CA.

\subsubsection{Software framework}
\begin{figure}[t]
  \begin{center}
    \resizebox*{8.2cm}{!}{\includegraphics{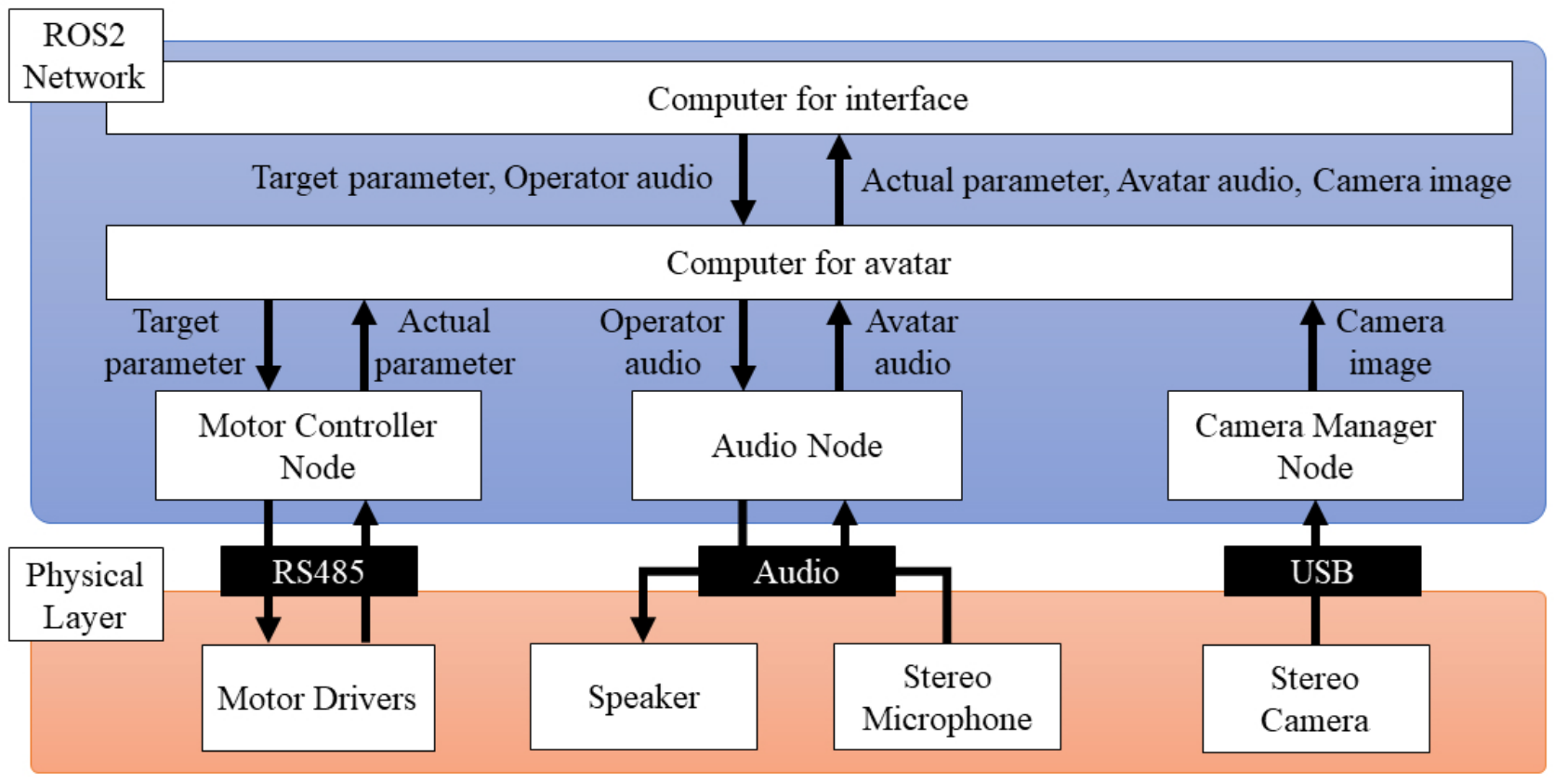}}
    \caption{Software framework of the avatar}
    \label{fig:software_framework}
  \end{center}
\end{figure}

The CA is driven by the system based on the ROS2 \cite{ros2}.
Fig.~\ref{fig:software_framework} shows the software framework.
In ROS2, each program unit operates independently and each unit exchanges data with each other.
The program unit in ROS is called the node, and the data exchanged during the nodes is called message.
Nodes can communicate with other nodes on the same network, even if they are running on different computers.
By using these functions, the communication between the computer for interface and it for avatar is conducted.
The framework for controlling avatar consists of three main nodes: the Motor Controller Node is used to exchange the current and target angles of the motor with the driver board, the Audio Node is used to send and receive audio data, and the Camera Node is used to send the camera image.
In addition to these nodes, the nodes for mediating communication with the interface computer are used, but we omit their description in this paper.
In addition to periodic behavior, each node also performs behavior resulting from the receipt of messages.

\begin{itemize}
    \item{
Camera node\\
The acquired camera image is sent as a message for each frame.
    }
    \item{
Audio node\\
The node acquires stereo audio from the audio device at regular intervals and sends it  to the interface computer as a message.
In addition, audio data from the interface computer is received and is played back sequentially through the avatar's speakers.
    }
    \item{
    Motor controller node\\
The node acquires the current angle, angular velocity, and input ratio from each driver board periodically, and sends them to the interface computer as a Joint State Message, which is a standard message type of ROS2.
The node also receives the target angle, angular velocity, and input limit of the motor from the interface computer as a same type message and controls each driver board.
The input ratio and input limit are parameters used to determine the voltage applied to the motor during motor control, and are specified in the range of zero to $1.0$.
    }    
\end{itemize}

\subsubsection{Motor control}
The motors are controlled by the PID controller embedded on the driver board. 
The driver board receives the target angle of each motor from the computer, and controls the motor's angle to track them.
The angle of each motor is measured by using a potentiometer connected to the motor shaft and the built-in ADC.
The movement of facial expressions are normalized, and the deformation from the neutral state is represented by values from zero to $1.0$.

The motor driver on the driver board controls input voltage by the PWM (Pulse Width Modulation) signal and switches between forward and reverse rotation by the digital signal for deciding the rotational direction.
The PWM signal takes a value from zero to $1.0$ and represents the ratio of ON time to the certain cycle.
For example, when the PWM signal is $0.5$, the ON time equals the OFF time of the motor driver.
The PWM signal is limited by the input limit received from the computer.
Let $\sigma$ be the input signal that determines the direction of rotation, $u$ be the PWM signal, $u_{\mathrm{lim}}$ be the input limit received from the computer, $w$ and $w_{d}$ be the current angle and target angle of the motor, respectively, and $e = w_{d} - w$ be the error between them.
Then, the PWM signal input is determined from the following equations.
\begin{eqnarray}
u^{\prime} &=& k_{p} e + k_{i} \int e~\mathrm{dt} + k_{d} \dot{e},\\
u &=& \begin{cases}
|u^{\prime}| & \left( -u_{\mathrm{lim}} < u < u_{\mathrm{lim}}\right)\\
u_{\mathrm{lim}} & |u^{\prime}| \geq u_{\mathrm{lim}}
\end{cases},\\
\sigma &=& \begin{cases}
    0 & \left( u^{\prime} \geq 0 \right)\\
    1 & \left( u^{\prime} < 0 \right)
\end{cases},
\end{eqnarray}
where, $k_{p}$, $k_{i}$ and $k_{d}$ are the arbitrary positive control gains and can be set to different values for each motor.
The motors rotate forward direction and reverse direction when $\sigma=0$ and $\sigma=1$,  respectively.

Due to the speed and capacity of serial communication, the communication cycle between the computer and the driver board is lower than the control cycle inside the driver board.
If the target angle is set in the communication cycle, the target angle changes in steps and the motor does not operate smoothly.
Therefore, the target angle for actual control is calculated by linear interpolation using the target data received from the computer.
Let the target angle and angular velocity received from the computer at time $t_{0}$ be $w^{\prime}_{d}$ and $v$, respectively, the target angle at time $t_{0}$ be $w_{d,0} = w_{d}(t_{0})$ and the time elapsed from time $t_{0}$ be $\delta t$.
The target angle $w_{d}(t)$ at time $t$ is given by 
\begin{eqnarray}
w_{d}(t) &=& 
\begin{cases}
    w_{d,0} + v \delta t & \left( w_{d,0} + v \delta t \leq w_{d}^{\prime}\right)\\
    w_{d}^{\prime} & \left( w_{d,0} + v \delta t > w_{d}^{\prime}\right)
\end{cases}.
\end{eqnarray}
By giving the target angular velocity considering the communication cycle $T$, the target angle of the motor changes smoothly.
The target angular velocity is set by 
\begin{eqnarray}
v = \frac{w_{d}^{\prime} - w_{d,0}}{T}.
\end{eqnarray}

\section{Operation Interface}
The CA is equipped with functions to recognize the surrounding environment and the interlocutors.
This information is presented to the operator through the Operation Interface.
Information such as the operator's voice, facial expressions, and head and eye movements are also transmitted and reproduced on the CA through the operation interface.
For reproducing and presenting information, the information must be converted appropriately.
This section introduces the conversion and presentation methods of this information.

\subsection{Hardware settings}
\begin{figure}[t]
  \begin{center}
    \resizebox*{8.2cm}{!}{\includegraphics{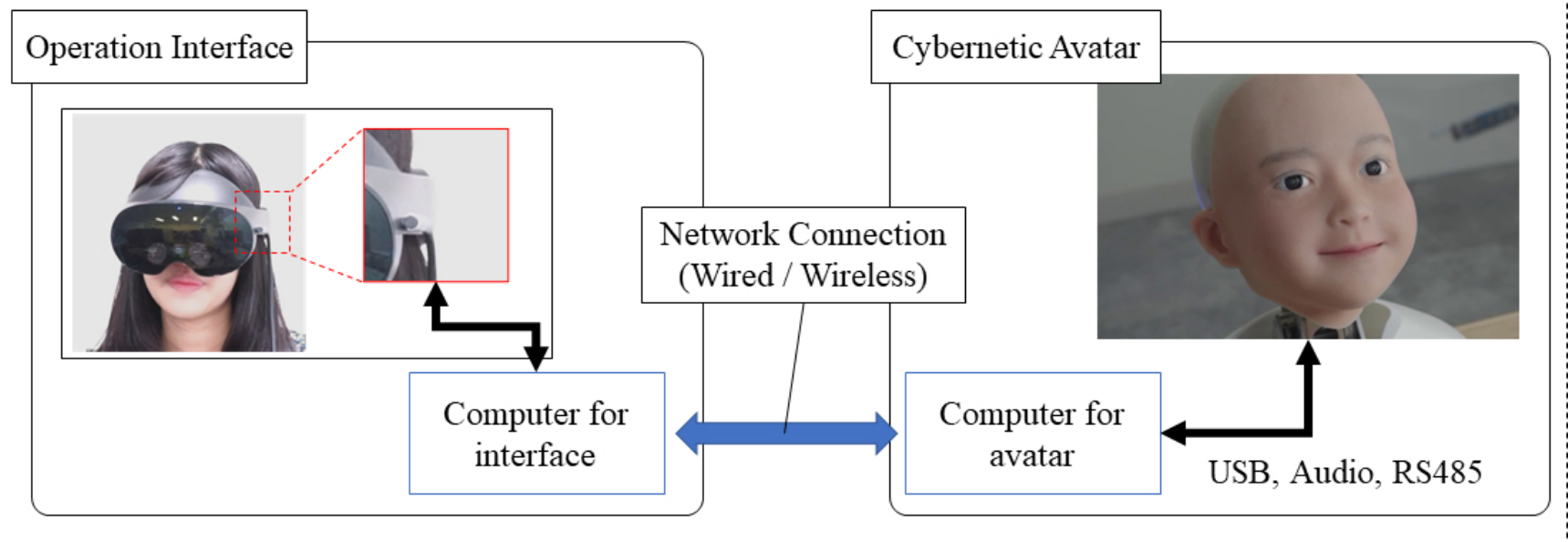}}
    \caption{Hardware settings}
    \label{fig:hardware_settings}
  \end{center}
\end{figure}

Fig.~\ref{fig:hardware_settings} shows the Hardware settings of the Operation Interface.
The Operation Interface consists of the HMD (Meta Quest Pro, Meta), a computer for communication with the CA and connection to the HMD.
This HMD uses an inside-out positional tracking method and acquires the position and posture of the HMD without using external sensors.
Therefore, the CA can be operated with a simple configuration of only the HMD and a computer for connection.
The HMD is equipped with an eye tracking unit and a facial tracking unit to acquire the operator's eye movements and facial expressions.
The position and orientation of the HMD are measured by the built-in sensors and multiple cameras mounted on the HMD.
The details of information that can be measured are referred to the official document \cite{face_tracking_meta}.
This paper introduces only the information used in this paper.

The position and orientation of the HMD are measured by using the cameras and IMU (Inertial Measurement Unit)  built in the HMD.
The neck motors are controlled based on the orientation of the HMD.
The eye tracking unit measures the direction of the sight as the vector.
The movement of the eyeball of the CA is controlled based on these vectors.
The facial tracking unit measures the motion of the lip, jaw, eyelids, and eyebrow.
The movement of the jaw motor is determined according to the degree of jaw opening, and the other facial motors are controlled according to corresponding expression data.
Note that the facial control points of the CA do not correspond exactly to the information that can be measured.
This is because the control points are designed based on the Action unit, which is more general information rather than the information that depends on each company's SDK.
In contrast, close motion can be achieved even when different HMDs are used by converting the information acquired by each HMD to the Action units.
This characteristic is significant in this research, which assumes no specified operators.
In this research, the motion of each motor is determined from multiple acquired information in order to absorb differences between the CA and the Operator's expressions as much as possible.
The details of the facial expression movements are explained in the following subsection.

\subsection{Software framework}
\begin{figure}[t]
  \begin{center}
    \resizebox*{8.2cm}{!}{\includegraphics{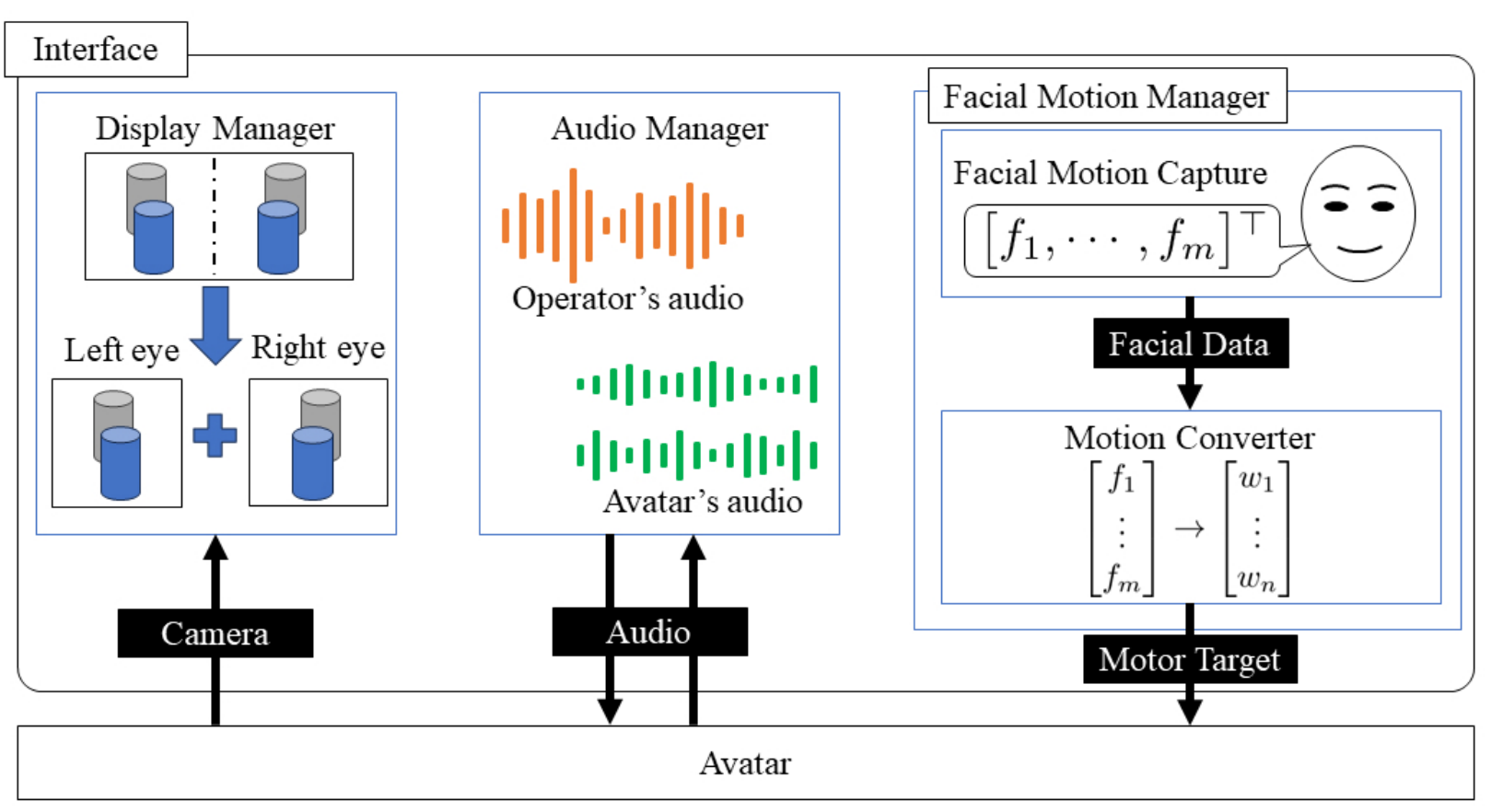}}
    \caption{Software framework of the interface}
    \label{fig:interface_framework}
  \end{center}
\end{figure}

Fig.~\ref{fig:interface_framework} shows the software framework of the Operation Interface.
The interface consists of three components: Display Manager for handling the HMD's screen, Audio Manager for audio processing, and Facial Motion Manager for converting the operator's facial expressions to Action units.
Display Manager renders the stereo camera image received from the CA to the left and right display of the HMD.
Audio Manager plays the stereo audio received from the CA on the HMD's speaker, and sends the operator's audio achieved by the microphone on the HMD to the CA.
Facial Motion Manager consists of the Facial Motion Capture, which obtains the information by the eye tracking unit and face tracking unit, and Motion Converter, which converts the operator's facial expression to the motion target of the facial motors.

There will be a slight discrepancy in the timing of audio and video reception and transmission.
In the exchange of audio data, audio data is received and transmitted between the Operation Interface and the CA at regular intervals.
Therefore, a certain amount of data is exchanged at regular intervals.
For example, if the Operation Interface communicates with the CA in 10 ms cycles, 10 ms data chunks are exchanged in each communication.
In addition, the delay in reception of camera image data is larger than that of audio data.
It takes time for the motor to reach the target value after the target is set.
These cause a gap between the avatar's facial expression and audio, and between the image and audio presented to the operator.
To eliminate this gap, a delay is introduced in the playback of audio data and video.
Adjusting the delay time eliminates the gap between the playback timing of the audio and the video, between the playback timing of the audio and facial expressions of the CA.
This method adjusts the delay to the one with the largest delay.
In this paper, the audio data transmitted to the CA is delayed according to the movement of the motor, and the audio data received from the CA is delayed according to the camera image.

\subsubsection{Presentation of Avatar's Perception}
\begin{figure}[t]
  \begin{center}
    \resizebox*{8.2cm}{!}{\includegraphics{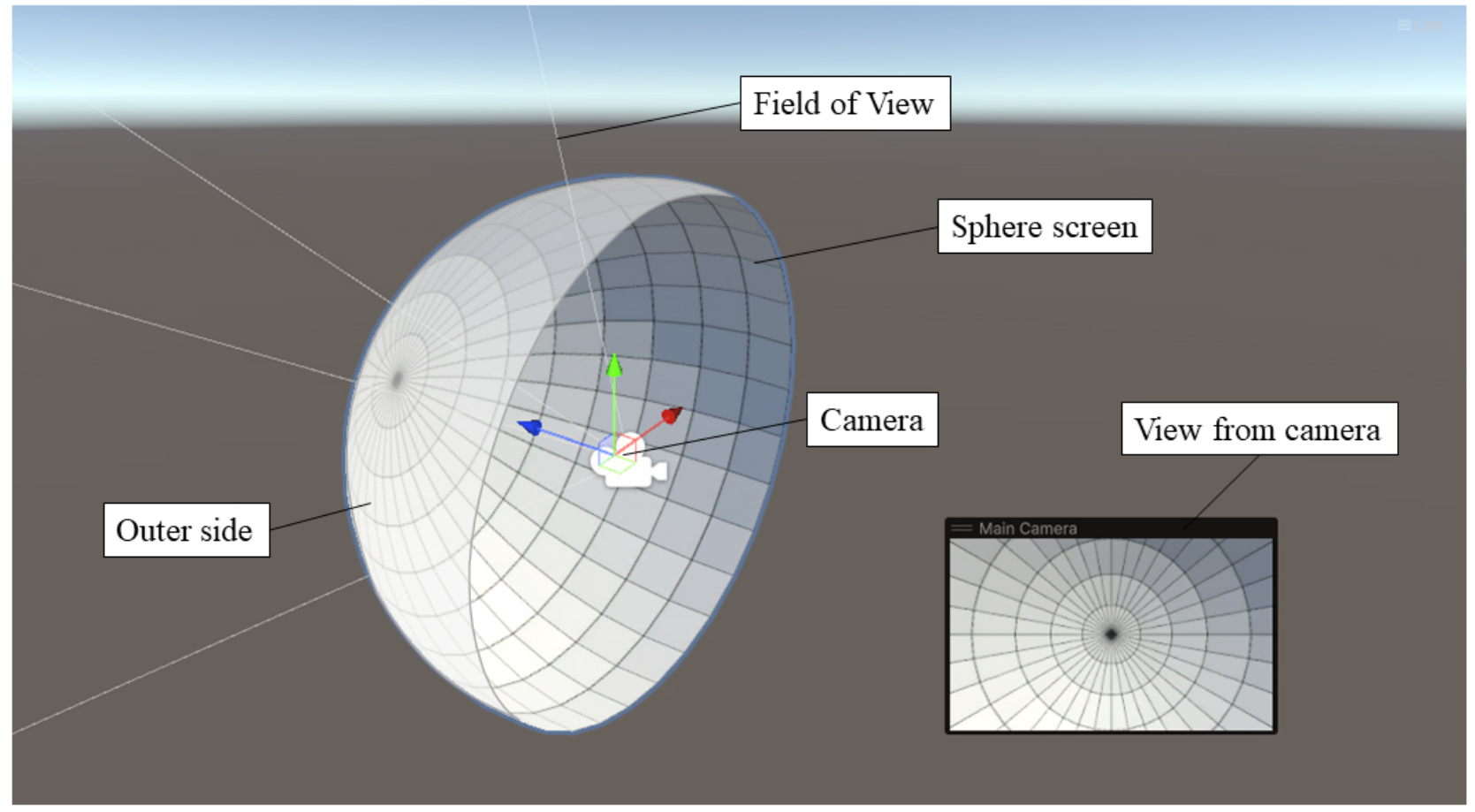}}
    \caption{Interface on Unity}
    \label{fig:hmd_interface}
  \end{center}
\end{figure}

\begin{figure}[t]
  \begin{center}
    \resizebox*{8.2cm}{!}{\includegraphics{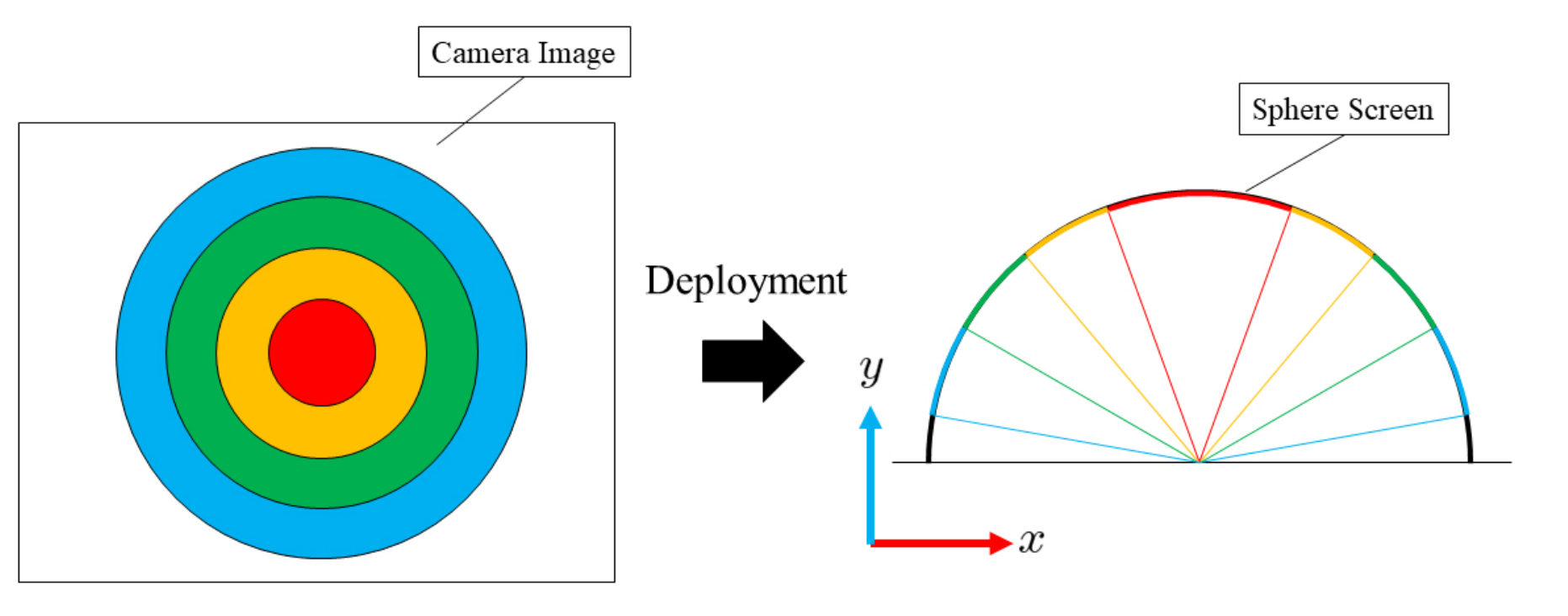}}
    \caption{Deployment of image to the screen}
    \label{fig:screen_map}
  \end{center}
\end{figure}

This section introduces how the camera image and audio information received from the CA is presented to the operator.
Fig.~\ref{fig:hmd_interface} shows the interface on Unity.
The images acquired by the CA's eye camera are transferred to the Operation Interface via the ROS2 network and projected onto a hemispherical screen on the Interface.
Originally, the backside of the hemispherical screen on the interface is transparent, but it is shown translucent in Fig.~\ref{fig:hmd_interface}.
Fig.~\ref{fig:screen_map} shows an overview of the projection of the camera image onto the screen.
The image is projected onto the screen so that the angle of view of the image acquired by the eye camera and the projected position on the screen roughly coincide.
However, it does not strictly consider the camera distortion.
The image viewed from the camera on Unity, that represents the operator's field of view, is projected onto the left and right displays on the HMD.
The images from the left and right eye cameras are projected onto the corresponding HMD left and right displays, respectively.
The center of the hemispherical display is aligned with the camera origin in Unity.
With the above settings, an image that is almost the same as the image seen from the eye camera is presented to the operator through the display on the HMD.
The Operation Interface also presents audio data transferred from the CA to the operator.
The audio data transferred from the CA is stereo audio data converted into binary array data.
This binary array is sequentially converted into stereo audio data and output from the left and right speakers of the HMD.
The audio acquired at the left and right ear of the CA are played on the left and right speaker of the HMD, respectively.

\subsubsection{Operator Movement Mapping}
This section introduces details of the process of transferring the operator's facial expressions, audio, and head and eye movements to the CA.
The operator's audio is transferred to the CA in the same manner as audio transfer from the CA to the Operator Interface.
The operator's head orientation, eye and facial expressions are converted into target motor angle by the Motion Converter and transferred to the CA via the ROS2 network.
By adjusting the parameter of the Motion Converter, various effects can be added to the facial expressions of the CA.
For example, conversion of the actual facial expressions into more exaggerated expressions may give an active impression to the interlocutor.
Investigating the effects of these transformations on the interlocutor is one of our future research issues.
In this study, the parameters were adjusted to reproduce the operator's facial expressions as well as possible on the CA, as a criterion for these future studies.
Let the values of each facital expression data be $f_{i}, (i=1,\cdots,m)$ and the target angle of each motor be $w_{i}, (i=1, \cdots, n)$.
The target angle of each motor is determined by 
\begin{eqnarray}
    \begin{bmatrix}
    w_{1}\\
    \vdots\\
    w_{n}
    \end{bmatrix}    
    =
    \bm{F}_{o} + 
    \bm{A}
    \begin{bmatrix}
    f_{1}\\
    \vdots\\
    f_{m}
    \end{bmatrix},
\end{eqnarray}
where, $\bm{F}_{o}\in \mathbb{R}^{n \times 1}$ is a constant that represents the motor target angle when all expression data are zero, and $\bm{A}\in\mathbb{R}^{m \times n}$ is the weight for converting each expression data to each motor target angle.
By adjusting $\bm{F}_{o}$ and $\bm{A}$ to match the operator's and CA's facial expressions as closely as possible, the operator's expression is reproduced on the CA.

\section{Experimental Verifications}
The experiments were conducted to verify the effectiveness of the proposed systems.
For verifying the CA, the CA was controlled based on predefined motions without an operator, and also controlled based on the operator's facial expression through the Operation Interface.
In addition, the view and audio presented to an operator through Operation interface were also verified.

\subsection{Facial expressions}
For verifying motions of the CA, we verified facial expressions of the CA when controlled based on predefined motions and controlled based on an operator's face expression.

\subsubsection{Predefined Expressions}
\begin{figure*}[t]
  \begin{center}
    \subfloat[Predefined expressions]{
    \resizebox*{8.2cm}{!}{\includegraphics{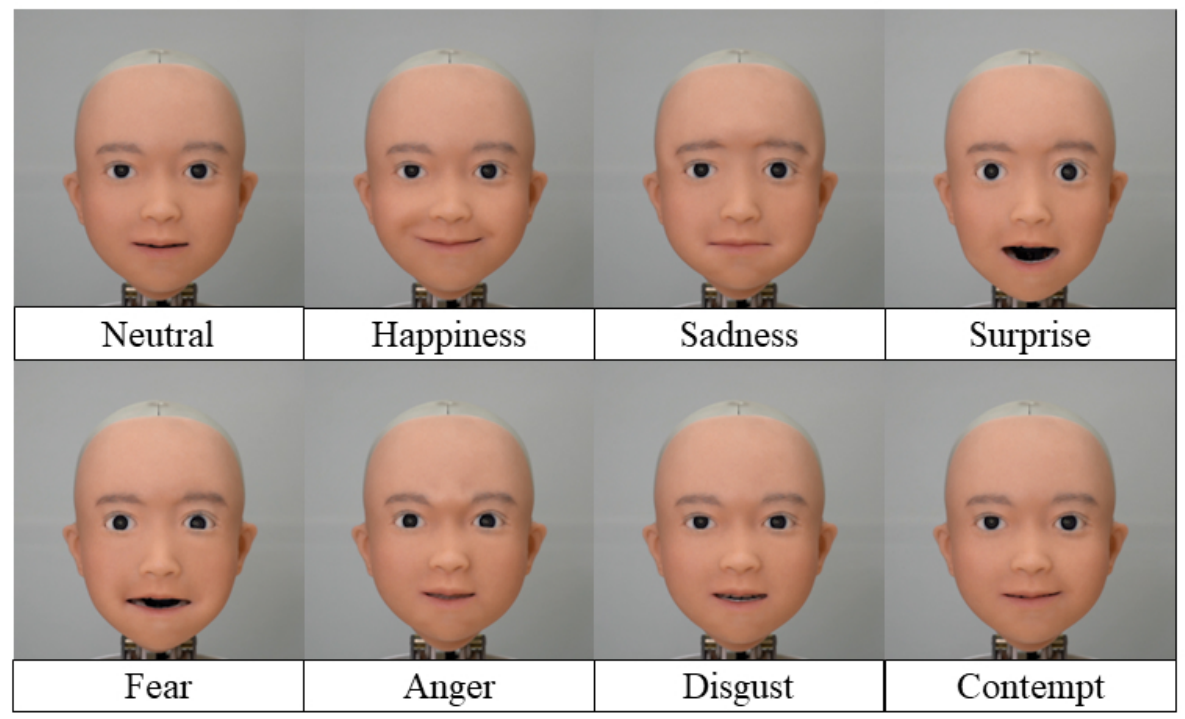}}    
    \label{fig:preset_emotion}}
    \subfloat[Operator-reflected expressions]{
    \resizebox*{8.2cm}{!}{\includegraphics{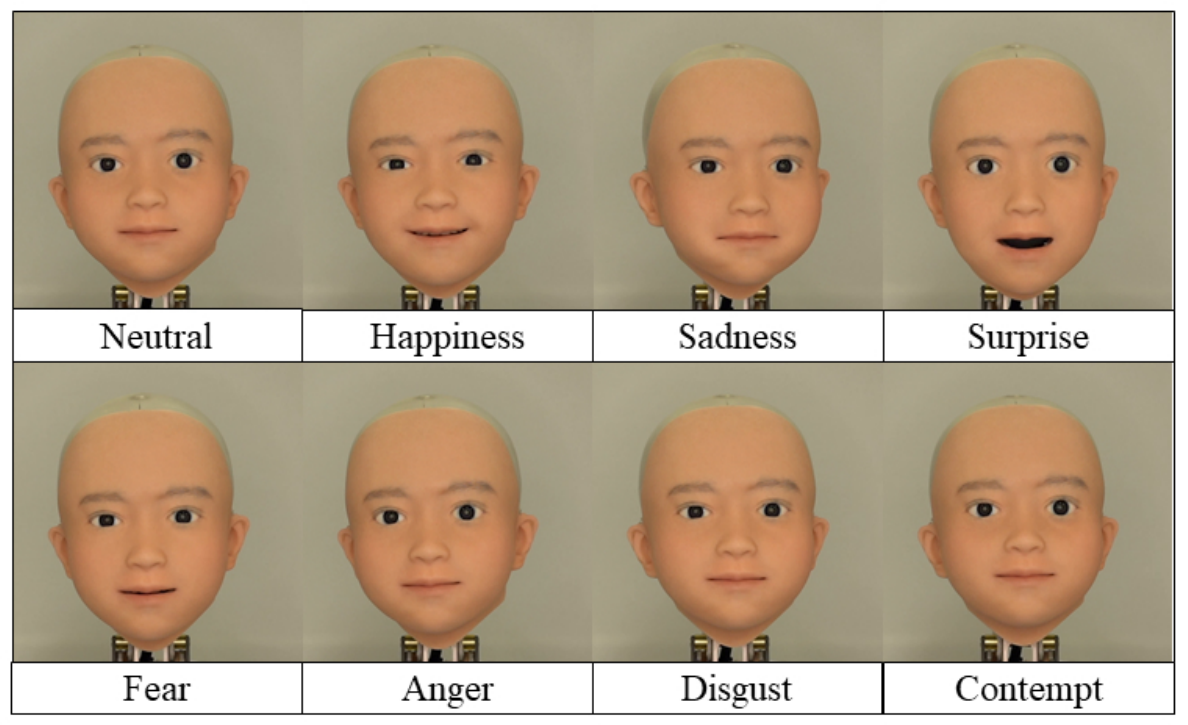}}    
    \label{fig:result_exp_face_if}}
    \caption{Result in the facial expressions experiment}
  \end{center}  
\end{figure*}

The facial expressions were expressed on the CA by controlling the facial actuator based on Table~\ref{tab:emotions_and_au}.
Fig.~\ref{fig:preset_emotion} shows the CA when expressing each facial expression,  and Table~\ref{tab:target_emote} shows the target values of the facial actuation points.
As shown in Fig.~\ref{fig:preset_emotion}, it was confirmed that the CA is capable of comprehensively the seven basic facial expressions.
The expression of disgust is sufficiently identifiable, although the facial expression changes less than the other expressions because no direct facial actuation point exists and close points are substituted.

\begin{table*}[t]
    \caption{Relationship between head motion and actuators}
    \centering
    \begin{tabular}{l|ccccccc}
         \toprule
         No. and  Description & Happiness & Sadness & Surprise & Fear & Anger & Disgust & Contempt\\ \hline
         4: Upper eyelid open      & -  & -   & 1.0   & 1.0   & 1.0   & -     & -     \\
         5: Upper eyelid close     & -  & -   & -     & -     & -     & -     & -     \\    
         6: Lower eyelid open      & -  & -   & -     & -     & -     & -     & -     \\
         7: Lower eyelid close     & -  & -   & -     & 0.7   & 0.5   & 0.3   & -     \\
         8: Left outer brow up     & -  & -   & 0.6   & 0.6   & -     & -     & -     \\
         9: Right outer brow up    & -  & -   & 0.6   & 0.6   & -     & -     & -     \\
         10: Left outer brow down   & -  & 1.0 & -     & -     & -     & -     & -     \\
         11: Right outer brow down  & -  & 1.0 & -     & -     & -     & -     & -     \\
         12: Left inner brow up    & -  & 1.0 & 0.6   & 0.6   & -     & -     & -     \\
         13: Right inner brow up   & -  & 1.0 & 0.6   & 0.6   & -     & -     & -     \\
         14: Left inner brow frown    & -  & -   & -     & -     & 1.0   & 0.6   & -     \\
         15: Right inner brow frown    & -  & -   & -     & -     & 1.0   & 0.6   & -     \\
         16: Left cheek pull & 1.0& -   & -     & -     & -     & -     & -     \\
         17: Right cheek pull & 1.0& -   & -     & -     & -     & -     & -     \\
         18: Upper lip up & -  & -   & 1.0   & 1.0   & 1.0   & 1.0   & -     \\
         19: Lower lip down & -  & -   & 1.0   & 1.0   & 1.0   & 1.0   & -     \\
         20: Left mouth corner pull & -  & -   & -     & 1.0   & -     & -     & 1.0   \\
         21: Right mouth corner pull & -  & -   & -     & 1.0   & -     & -     & -     \\
         22: Mouth pucker & -  & -   & -     & -     & 0.6   & -     & -     \\
         23: Left mouth corner up& 1.0& -   & -     & -     & -     & -     & 1.0   \\
         24: Right mouth corner up    & 1.0& -   & -     & -     & -     & -     & -     \\
         25: Left mouth corner down  & -  & 1.0 & -     & 1.0   & -     & -     & -     \\
         26: Right mouth corner down    & -  & 1.0 & -     & 1.0   & -     & -     & -     \\
         27: Jaw open    & -  & - & 1.0   & 0.8   & -   & -     & -     \\ 
         28: Jaw close    & -  & 1.0 & -   & -  & 1.0   & -     & -     \\ \hline         
    \end{tabular}    
    \label{tab:target_emote}
\end{table*}

\subsubsection{Operator-reflected Expressions}
We verified the facial expressions when the operator controlled the face with the Operation interface.
The operator changed his own facial expression to be similar to the expression shown in Fig.~\ref{fig:preset_emotion}.
Fig.~\ref{fig:result_exp_face_if} shows the results of the verifications.
As shown in Fig.~\ref{fig:result_exp_face_if}, Happiness, Sadness, and Surprise can be discriminated against although the changes in facial expressions are small.
In addition, operator's behaviors, such as eye and eyelid movements which did not appear in the pre-designed motions, were reproduced.
In contrast, the facial movements were smaller than the pre-designed expressions.
Therefore, facial expressions such as Fear and Anger, which express emotion with large facial movements, were not clearly expressed.
One of the reasons for this may be that it is difficult for operators to move the his own face to the maximum extent when they control CA through the Operation interface.
As shown in Table~\ref{tab:target_emote}, the amount of motion of each facial point is large when the CA is operated with the pre-designed motion.
This suggests that the CA's facial movement will be small when the control parameters are set to move the CA's facial expressions so that they are similar to those of the operator.

\subsection{Operation Interface}
We confirmed the information presented by the Operation interface.
The video and audio perceptions acquired by the camera and microphone mounted on the CA were confirmed.

\subsubsection{Visual perception}
\begin{figure}[t]
  \begin{center}
    \resizebox*{8.2cm}{!}{\includegraphics{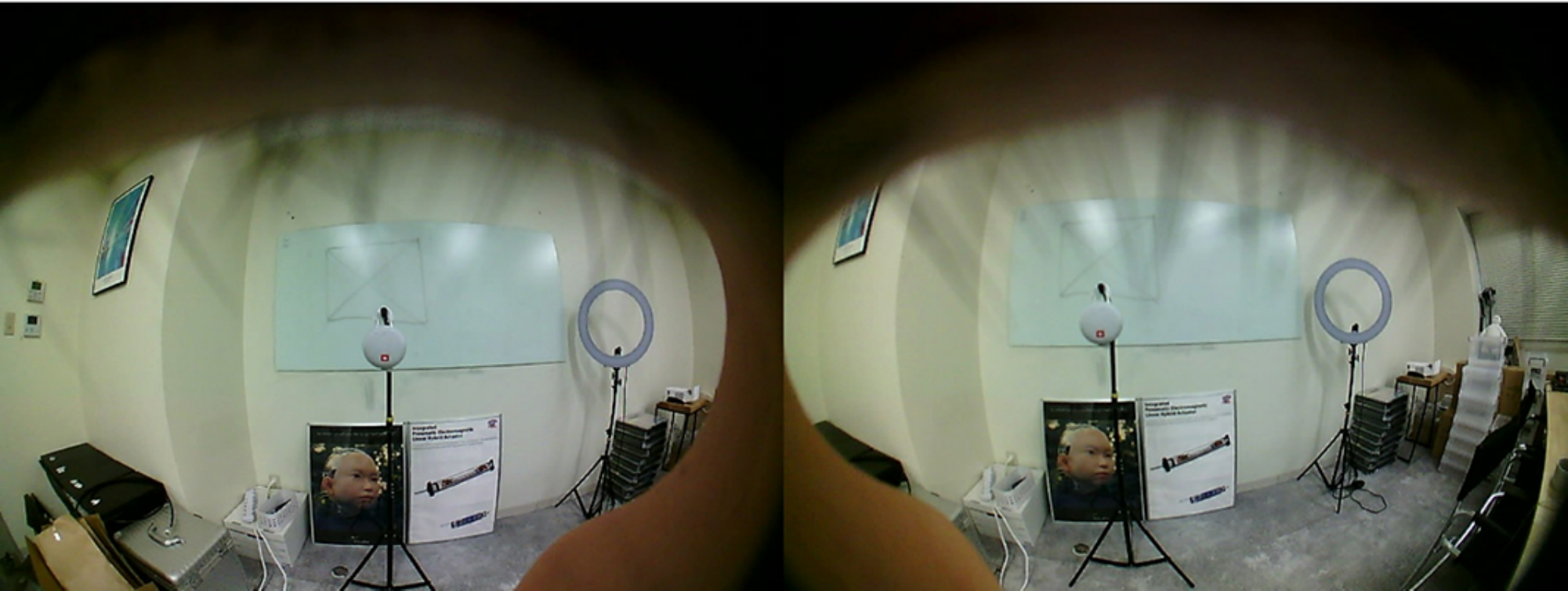}}
    \caption{Stereo camera image in the experiments}
    \label{fig:result_exp_eye}
  \end{center}
\end{figure}

The visual perception for the operator through the Operation Interface was confirmed.
Fig.~\ref{fig:result_exp_eye} shows images of cameras with both eyes when an object was placed in front of the avatar.
A portion of this image is attached to a spherical screen when presented to the operator.
Fig.~\ref{fig:result_exp_eye} shows that the relative relationship between the object in front of the avatar and the wall behind it is different in the left and right camera images.
It is expected that the operator is able to recognize the depth of the object and the positional relationship between them from this parallax image.

\subsubsection{Audio perception}
\begin{figure}[t]
  \begin{center}
    \subfloat[Settings of audio experiments]{
    \resizebox*{8.2cm}{!}{\includegraphics{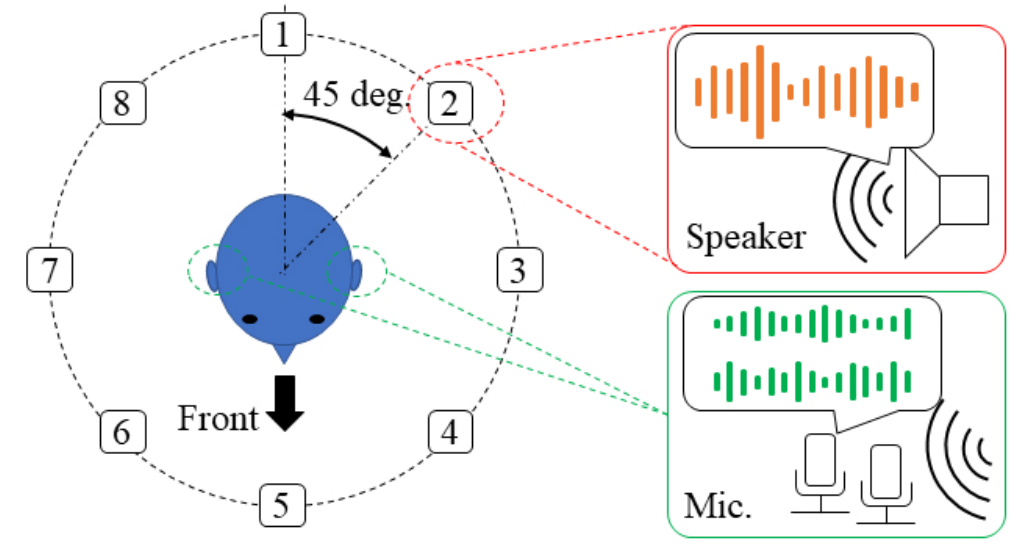}}    
    \label{fig:env_exp_audio}}
    \\
    \subfloat[Result in the audio experiment]{
    \resizebox*{8.2cm}{!}{\includegraphics{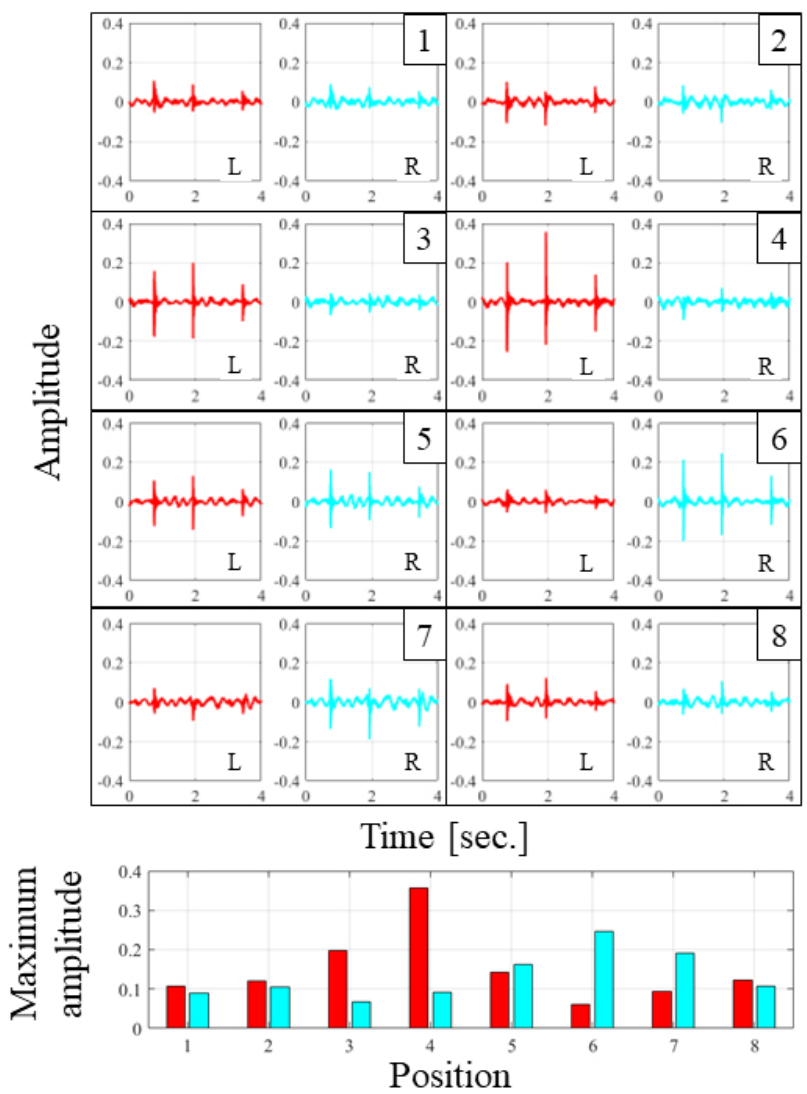}}    
    \label{fig:result_exp_audio}}
    \caption{Audio verification Experiment}
    \label{fig:audio_exp}
  \end{center}
\end{figure}

We conducted the verification experiment for audio perception presented to the operator by Operation Interface.
As shown in Fig.~\ref{fig:env_exp_audio}, the same audio from different locations on the circumference of the circle centered at the avatar was outputted, and the differences in the audio data when the avatar heard it were verified.
The radius of the circle was set to 1.0 m.
Fig.~\ref{fig:result_exp_audio} shows the waveform data and the maximum amplitude of the audio data which the avatar heard.
Fig.~\ref{fig:result_exp_audio} shows the significant difference in the amplitude of the left and right audio data in the cases where audio is outputted from the left and right sides of the front (Positions 3, 4, 6, and 7).
In these cases, the amplitude of the side closest to the sound source is larger and it of the opposite side  is smaller.
In contrast, there was no significant difference in the audio data between left and right, in the cases where audio was outputted from the front side or rear side (Positions 1 and 5).
In addition, in the cases where audio is outputted from the left and right sides of the rear (Positions 2 and 8), there was no significant difference in the audio data between left and right and the volume of the audio was small.
We consider that these differences were due to the fact that the auricle on the avatar collects sounds from the front side and blocks sounds from the rear side.
The above results indicate the possibility of the operator to estimate the direction of the sound source by the proposed system.

\subsection{Operation while communication}
\begin{figure}[t]
  \begin{center}
    \resizebox*{8.2cm}{!}{\includegraphics{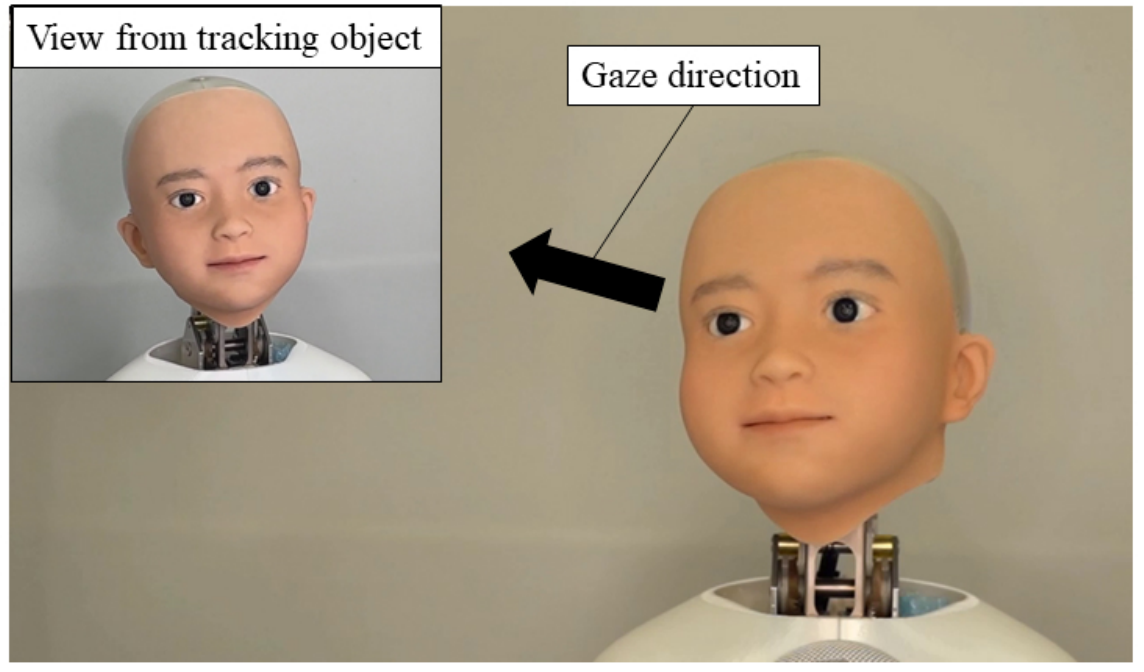}}
    \caption{Result in eye and head motion experiment}
    \label{fig:gaze_direction}
  \end{center}
\end{figure}

\begin{figure}[t]
  \begin{center}
    \resizebox*{8.2cm}{!}{\includegraphics{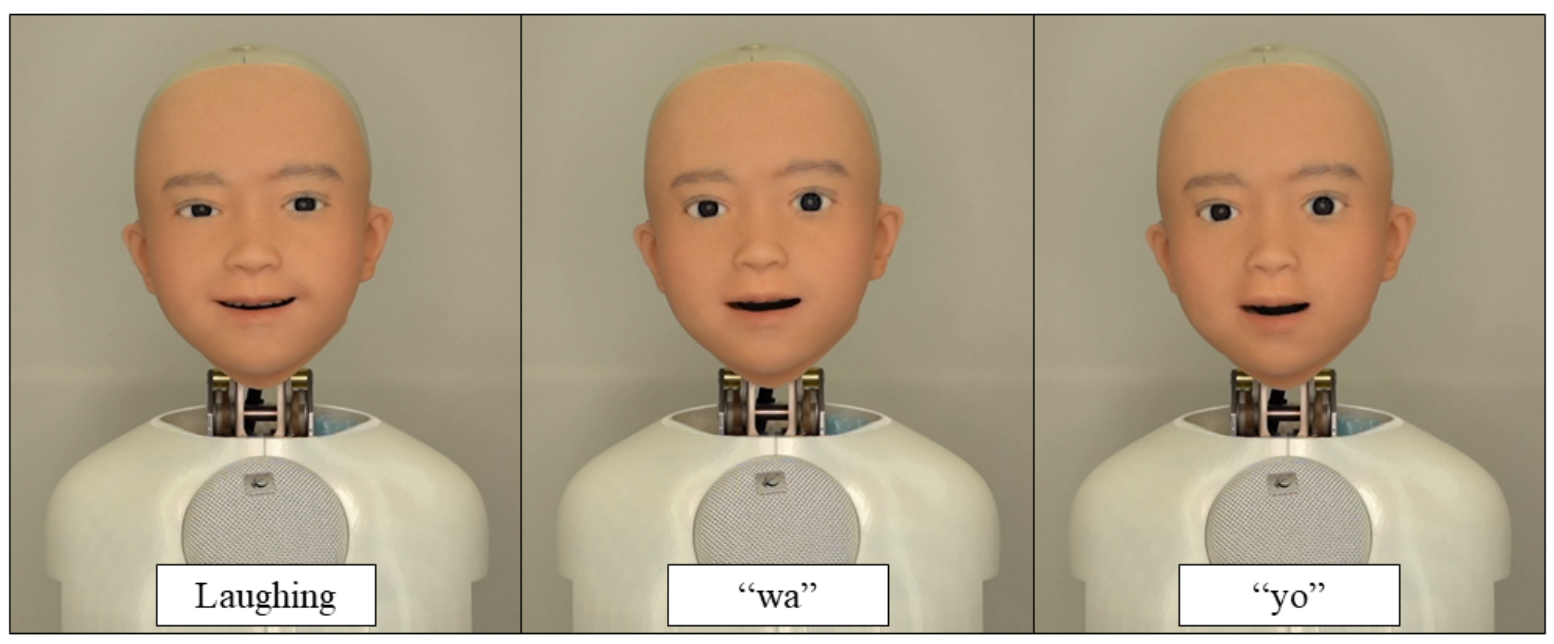}}
    \caption{Result in greeting experiment}
    \label{fig:exp_greet}
  \end{center}
\end{figure}

We conducted the simple experimental validation of the communication through the CA and Operation Interface.
The experimental video is attached as an appendix.
Following verifications were conducted.
\begin{enumerate}
    \item Following object with the eyes and neck movement
    \item Simple greeting through the Operation Interface
\end{enumerate}
In verification 1, an object was moved in front of the CA, and the operator followed it through the Operation Interface.
In verification 2, the operator made a brief greeting through the Operation Interface.
Fig.~\ref{fig:gaze_direction}--\ref{fig:exp_greet} show the images of each verification cropped from the video.
The neck and eyes moved simultaneously to see objects at the edge of the field of view, as shown in Fig.~\ref{fig:gaze_direction}.
This result means that the proposed system is able to see a wider area than systems that move only the neck without eyes.
In addition, the system has the  possibility of reproducing eye contact, averted eyes, and looking up movements by eye movements.
Movements of the jaw, corners of the mouth, cheeks, and blinking can be observed from Fig.~\ref{fig:exp_greet}.
The CA reproduced the operator's non-verbal behaviors, which are difficult to express in cases where mouth movements are generated from voices.

\section{Discussion}
The experimental verifications confirmed that the developed CA `Yui' can express emotions such as happiness, surprise, and anger.
By using more abundant facial motion than the electric driven android `ibuki' \cite{ibuki} developed by the authors, the CA expressed the seven basic emotions: Happiness, Sadness, Surprise, Fear, Anger, Disgust, and Contempt.
Pneumatic-driven androids with a larger degree of freedom than the developed CA have also been developed \cite{Nikola}, but their size and drive system make it difficult to utilize them as a mobile avatar.
The electric driven android EveR-4 H33 \cite{ever4_h33} has a larger degree of freedom than the developed CA.
However, this android does not have the components for perceptual functions such as microphone or camera, and are not developed as a remote control avatar.
The developed CA achieves a large degree of freedom with a small number of actuators by controlling different degrees of freedom in the forward and reverse rotation of the motor.
This structure realizes miniaturization of the CA and the reproducing rich expressions on the CA.
In addition, more detailed expressions are realized compared to the simplified expression method \cite{icub3} because the deformation of the face changes depending on the pull amount of the wire. 
Rich facial expressions on the CA may allow the operator to express rich emotions to the interlocutor via the CA.
In addition, it showed the possibility of presenting the direction of the sound source and the perspective and relative relationships of the objects to the operator through stereo audio and stereo images.
In the verification of the stereo audio, it was confirmed that there is a difference in the audio waveform acquired from the microphones mounted on the left and right ears depending on the direction of the sound source.
These features enable the operator to perceive the positional relationship with the interlocutor, and is expected to improve the interaction experience between the operator and the interlocutor through the CA.
These results suggest that communication through the developed system enables the exchange of more non-verbal information than in previous studies.
Therefore, the developed system may contribute to the realization of direct communication between operators and interlocutors through CA.

In contrast, in the experiment of reproducing the operator's facial expressions on the CA, deformations of the facial expressions were smaller than those given in the pre-designed expressions.
Some challenges remain in the method of reproducing the operator's facial expressions on the CA.
In addition, the avatar and the operator's facial expressions were synchronized through trial and error pre-tuning of parameters for conversion.
When the system is used by a large number of operators, a mechanism to smoothly or adaptively adjust  these parameters is necessary.

Future issues include the study of a mechanism to adjust the synchronization parameters between the operator and the avatar, the implementation of arms and movement mechanisms, and the comparison between communication using the proposed system and face to face communication.
In addition, discussion about the necessity of accurately reproducing the operator's facial expressions to the CA is needed.
For example, if the deformation of the operator's facial expression is small, there is a possibility of improving the quality of the communication by amplifying the facial expressions of the operator.
The movement of the mouth when speaking varies from person to person, and some people speak almost without moving their mouths.
In such cases, reproducing the operator's mouth movements on the CA may cause a strong discomfort.
By adding additional effects to the original expression according to the use case, the authors believe that it is possible to go beyond face-to-face communication in the future.
In addition, communication between people is greatly affected not only by facial expressions but also by the exchange of physical information such as hand movements and distance from the person \cite{Mehrabian2017-tz}.
Therefore, with the aim of achieving more interactive communication, we will also work on the implementation of the hands and arms and a mobilization mechanism.
In the comparative verification of the proposed system, we will investigate in more detail how it differs from face-to-face communication, focusing on the presence felt by the operator and interlocutor via Operation Interface, and the avatar's impression to the interlocutor.

\section{Conclusion}
In this paper, we developed a head unit of a new remote communication cybernetic avatar  `Yui' and an immersive interface to operate it.
The developed head unit can express much richer facial expressions than previously developed mobile avatars.
The operator can obtain stereoscopic sound and vision through the immersive interface, which provides a stronger sense of presence compared to non-immersive interfaces such as displays.
In addition, it is suggested that the operator's facial expression could be reproduced on the avatar by controlling actuators based on the operator's facial expression and eye movement acquired by the HMD's sensors.
The developed system can improve the communication experience for both the operator and the interlocutor.
The developed system is expected to make a significant contribution to the realization of direct communication between the operator and the interlocutor through avatars.
Future prospects include the investigation of the effects on the operator and interlocutor by using the developed avatar and interface, and the development of a full body for use as a movable avatar.

\section*{Other Material}
Video: https://youtu.be/D0R2R-64RKU

\bibliographystyle{ieeetr}
\bibliography{bibref}

\EOD

\end{document}